\documentclass[10pt,twocolumn,letterpaper]{article}
\usepackage[pagenumbers]{cvpr}            
\usepackage{graphicx}
\usepackage{amsmath}
\usepackage{amssymb}
\usepackage{booktabs}
\usepackage{multirow}
\usepackage{xcolor}
\usepackage{bm}
\usepackage[normalem]{ulem}
\usepackage{wrapfig}
\usepackage{soul}
\usepackage[accsupp]{axessibility}  

\newcommand{\you}[1]{{#1}}
\newcommand{\huiqi}[1]{{#1}}
\newcommand{\AY}[1]{}
\newcommand{\nt}[1]{{#1}}

\newcommand{\youRevision}[1]{{#1}}
\newcommand{\huiqiRevision}[1]{{#1}}

\usepackage{caption}
\usepackage{algorithm}
\usepackage{algpseudocode}
\newcommand{\myrefeq}[1]{Equation~\ref{#1}}
\newcommand{\myreffig}[1]{Figure~\ref{#1}}
\newcommand{\myreftab}[1]{Table~\ref{#1}}
\newcommand{\myrefsec}[1]{\secref{#1}}

\usepackage{amsmath,amsfonts,bm}

\def\secref#1{section~\ref{#1}}
\def\eqref#1{equation~\ref{#1}}
\def\1{\bm{1}}

\DeclareMathAlphabet{\mathsfit}{\encodingdefault}{\sfdefault}{m}{sl}
\SetMathAlphabet{\mathsfit}{bold}{\encodingdefault}{\sfdefault}{bx}{n}

\DeclareMathOperator*{\argmin}{arg\,min}

\newcommand{\Lapp}{\mathcal{L}_{\text{app}}}
\newcommand{\Ltemp}{\mathcal{L}_{\text{temp}}}

\usepackage{xspace}
\renewcommand*{\eg}{e.g.\@\xspace}
\renewcommand*{\ie}{i.e.\@\xspace}

\makeatletter
\renewcommand*{\etc}{
    \@ifnextchar{.}
        {etc}%
        {etc.\@\xspace}
}
\makeatother
\usepackage[pagebackref,breaklinks,colorlinks]{hyperref}

\usepackage[capitalize]{cleveref}
\crefname{section}{Sec.}{Secs.}
\Crefname{section}{Section}{Sections}
\Crefname{table}{Table}{Tables}
\crefname{table}{Tab.}{Tabs.}

\def\cvprPaperID{3480}
\def\confName{CVPR}
\def\confYear{2022}

\begin{document}

\title{TemporalUV: Capturing Loose Clothing with \\
Temporally Coherent UV Coordinates}

\author{You Xie\textsuperscript{$\dagger$}, Huiqi Mao\textsuperscript{$\ddagger$}, Angela Yao\textsuperscript{$\ddagger$}, Nils Thuerey\textsuperscript{$\dagger$}\\
\textsuperscript{$\dagger$}{Department of Informatics, Technical University of Munich}\\
\textsuperscript{$\ddagger$}{Department of Computer Science, National University of Singapore}\\
{\tt\small \{you.xie, nils.thuerey\}@tum.de, huiqi.mao@u.nus.edu, ayao@comp.nus.edu.sg}
}
\maketitle
\begin{abstract}
   We propose a novel approach to generate temporally coherent  UV coordinates for loose clothing. Our method is not constrained by human body outlines and can capture loose garments and hair. 
   We implemented a differentiable pipeline to learn UV mapping between a sequence of RGB inputs and textures via UV coordinates. 
   Instead of treating the UV coordinates of each frame separately, our data generation approach connects all UV coordinates via feature matching for temporal stability.
   Subsequently, a generative model is trained to balance the spatial quality and temporal stability. It is driven by supervised and unsupervised losses in both UV and image spaces.
    Our experiments show that the trained models output high-quality UV coordinates and generalize to new poses. 
    Once a sequence of UV coordinates has been inferred by our model, it can be used to flexibly synthesize new looks and modified visual styles.
    Compared to existing methods, our approach reduces the computational workload to animate new outfits by several orders of magnitude. 
\end{abstract}

\section{Introduction}
\label{sec:intro}
In image or video generation tasks~\cite{yang2018pose,tang2020xinggan} that involve people, it is crucial to obtain accurate representations of the 3D human shape and appearance to efficiently generate modified content.
In this context, \textit{UV coordinates} are a popular 2D representation that establish dense correspondences between 2D images and 3D surface-based representations of the human body. 
UV coordinates go beyond skeleton landmarks to encode human pose and shape,  
and are widely used in image/video editing, augmented reality, and human-computer interaction ~\cite{deng2018uv,farras2021rgb,gecer2021ostec}. 
{In this paper, we tackle video generation of people, with a focus on efficiency and capturing loose clothing. Unlike previous works \cite{vondrick2016generating, saito2017temporal,wang2020imaginator} which use large networks to capture motion and appearance, we train a model to generate temporally coherent UV coordinates. We use a single, fixed texture to store appearance information so that our model can solely focus on learning UV dynamics.}

Human body UV coordinates 
can be derived indirectly from estimates of 3D shape models~\cite{pavlakos2018learning,bogo2016keep,madadi2018smplr,kocabas2020vibe} like SMPL~\cite{loper2015smpl}. 
Alternatively, direct estimation methods like DensePose~\cite{alp2018densepose} and UltraPose~\cite{yan2021ultrapose} 
bypass intermediate 3D models to directly output UV coordinates from a single RGB image. The convenience of direct methods has led to DensePose being widely used in animation and editing applications~\cite{neverova2018dense,neverova2019slim,ma2020unselfie,zhu2020simpose}.
Nevertheless, the UV coordinates obtained from SMPL and DensePose approximate only human body silhouettes in tight clothing. They do not capture loose clothing, such as long skirts or wide pants (see comparisons in \myreffig{fig:more_optimization} and \ref{fig:more_optimization_smpl}).
In addition, the methods for UV estimation 
work only on individual images. For video inputs, they 
are applied frame-by-frame \cite{pumarola2019unsupervised,zablotskaia2019dwnet} without considering the temporal relationship between frames. 
As such, the UV coordinates are inconsistent over time, so any re-targeted sequences will shift and jitter. 

In this paper, we focus on improving the spatial coverage and temporal coherence of UV coordinates generated from a sequence of 2D images. 
We target the ability to retain the full body plus clothing silhouette for arbitrary styles of clothing. Our approach is agnostic to the UV source, which we demonstrate via inputs from both DensePose~\cite{alp2018densepose} and SMPL model estimates~\cite{kocabas2020vibe}. \youRevision{For temporal coherence, we aim at achieving the point-to-point correspondences among different frames via UV coordinate maps, so that video sequences can be generated with one fixed texture.}

\begin{figure*}
\vspace{-2mm}
	\begin{center}
		\includegraphics[width=0.9\linewidth]{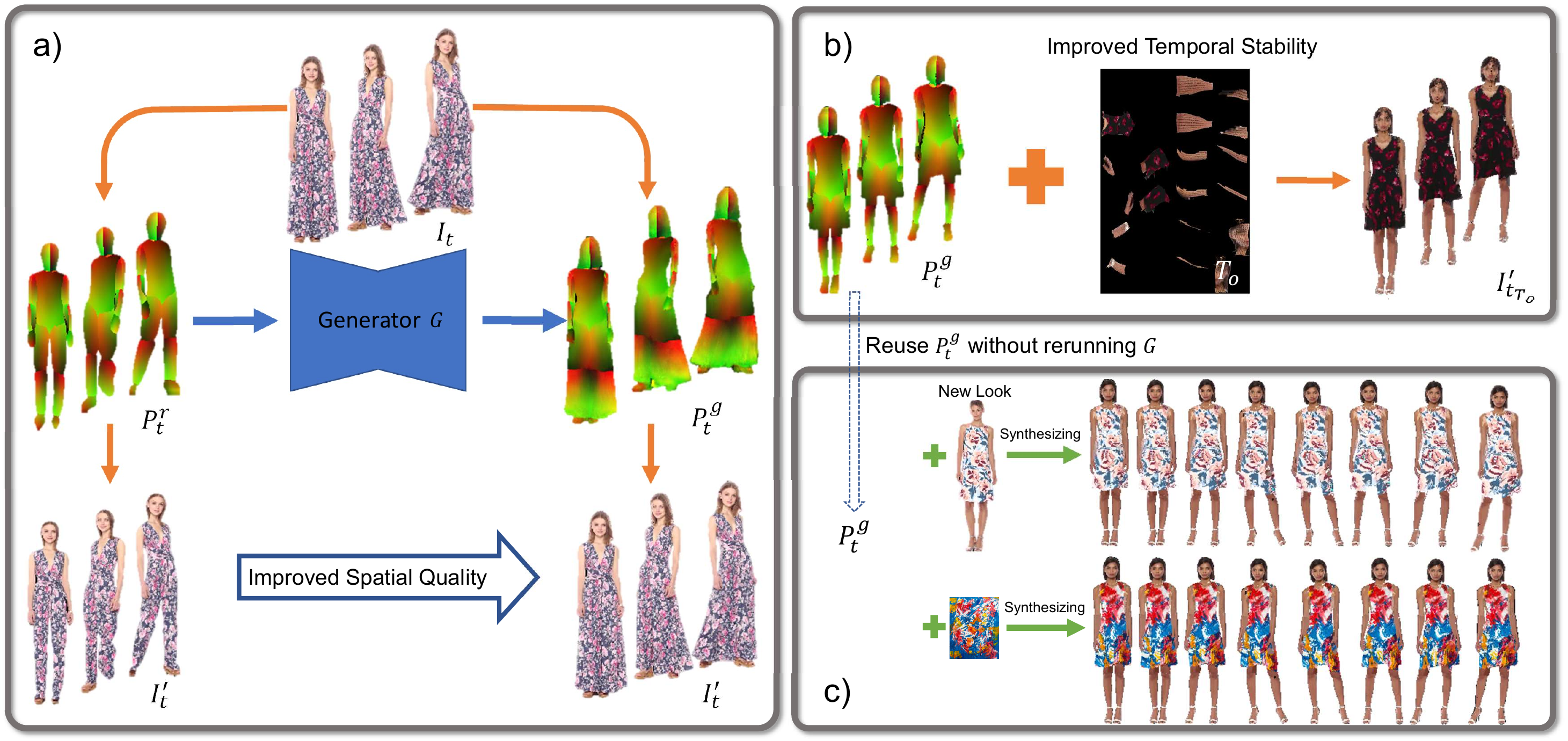}
	\end{center}
	\vspace{-5mm}
	\caption{a) Our method generates temporally coherent UV coordinates that capture loose clothing from off-the-shelf human pose UV estimates such as SMPL and DensePose~\cite{loper2015smpl,alp2018densepose}. 
	b) Generated 
	UV coordinates allow us to recover entire sequences from a constant texture map. 
	c) Virtual try-on and modifications of the look 
	can be easily achieved 
	with minimal computation via a simple lookup. 
	}
	\vspace{-4mm}
	\label{fig:teaser}
\end{figure*}

A core challenge of learning a model for extended and temporally coherent UV coordinates lies in the lack of data for direct supervision. 
Hence, we propose a novel learning scheme that combines both supervised and unsupervised components. We first pre-process a sequence of UV coordinates obtained from DensePose or SMPL via spatial extension and temporal stabilization 
to obtain 
training data for an initial training stage.
We then shift the learning gradually from supervised, with the pre-processed data, to unsupervised, 
driven by a differentiable UV mapping pipeline between the texture and image space. 

Our results demonstrate that using loss terms formulated in both UV and image space are crucial for generating high-quality UV coordinates with temporal coherence. As our generator does not take RGB images as input, the UV coordinates generated from our trained model can be directly paired with different texture maps to generate virtual try-on videos with a very simple lookup step. This is device-independent and orders of magnitude more efficient than other methods, which generate video outputs by evaluating neural networks. 
To summarize, our main contributions are
\vspace{-6mm}
\begin{itemize}
\setlength\itemsep{0.2em}
\item a model-agnostic method to extend UV coordinates 
to capture the complete appearance of the human body,
\item an approach to train neural networks that generate completed and temporally coherent UV coordinates without the need for ground truth, and
\item a highly efficient way to generate virtual try-on videos with arbitrary clothing styles and textures.
\end{itemize}
\vspace{-4mm}

\begin{figure*}[ht]
\vspace{-2mm}
	\begin{center}
		\includegraphics[width=0.75\linewidth]{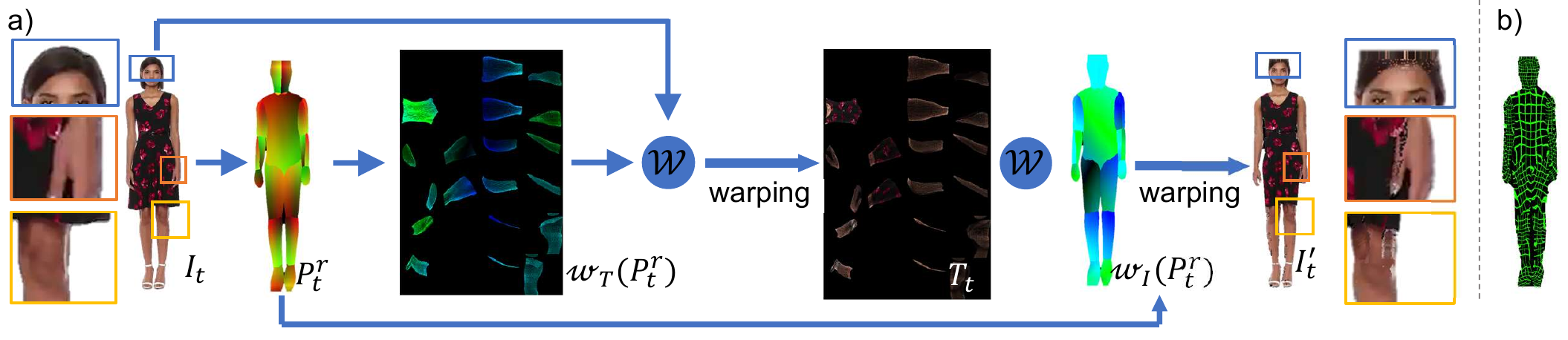}
	\end{center}
    \vspace{-5mm}
	\caption{a) Example mapping from $I_t$ to $T_t$ via $P^r_t$, and back to $I'_t$. $P^r_t$ cannot fully recover the image, and misses skirt, hair, and shoulder parts. Besides, colours inside the human body are also partially incorrect. b) Mapping results of $P^r_t$ with $T_{grid}$ as input. Most quadrants are preserved, indicating that the corresponding features are not destroyed after the UV mapping.
	}
	\label{fig:I2T2I}
	\vspace{-4mm}
\end{figure*}
\section{Related work}
\paragraph{Pose-guided generation.} 
Pose-guided methods~\cite{ma2017pose, balakrishnan2018synthesizing, siarohin2018deformable, pumarola2018unsupervised, neverova2018dense, grigorev2018coordinate, liu2019liquid} generate images of a person with designated target poses. To achieve realistic and high-quality generations, most methods~\cite{grigorev2018coordinate, liu2019liquid, neverova2018dense} tend to work with dense targets with 3D shape or surface models. Specifically, these methods rely on UV coordinates generated either from estimated SMPL model parameters~\cite{loper2015smpl} or directly via DensePose~\cite{alp2018densepose}. Neither the SMPL model nor DensePose is good at dealing with loose clothing, such as dresses. In this paper, we also work with %dense mappings like 
UV coordinates, though our pipeline focuses on \emph{improving} the quality of the raw UV coordinates from SMPL and DensePose to take on loose clothing.
Pose-guided video generation, also known as human motion transfer, generate videos based on a sequence of target poses~\cite{siarohin2019animating, zablotskaia2019dwnet, siarohin2019first, dong2019fw}. The appearance information is sourced from either images (image-to-video~\cite{siarohin2019animating,zablotskaia2019dwnet,yoon2021pose,siarohin2019first}) or videos (video-to-video~\cite{aberman2019deep, chan2019everybody,cheng2019multi,liu2019neural}).  

\vspace{-2mm}
\paragraph{Image-to-video.} 
An early example is MonkeyNet~\cite{siarohin2019animating}. While Monkeynet decouples appearance and motion information, it uses keypoints, 
which is insufficient for 
high-quality capture of human body or clothing with complex textures. We use DensePose UV coordinates as pose representation to improve this problem.

Closely related to our work is DwNet~\cite{zablotskaia2019dwnet}, which also uses DensePose UV coordinates as inputs. DwNet applies an encoder-decoder architecture that warps the human body from source to target poses. However, DwNet can be difficult to train due to its use of highly non-linear warping grids. The generator also needs to be re-run
for computing the warping grid each time the source image changes. 
Instead of predicting warping grids, our method works directly on the UV coordinates, making it independent of the source images. 
Once the target sequence of UV coordinates is generated, it can be applied for different textures without re-running the model regardless of complexity.

\vspace{-2mm}
\paragraph{Video-to-video.} These methods~\cite{aberman2019deep, chan2019everybody,cheng2019multi,liu2019neural}  have access to a source video and can therefore create richer models of the source subject than single image sources. 
In particular,~\cite{cheng2019multi} generates videos with spatial transformation of target poses, allowing it to capture loose clothing. However, all these works rely on 2D keypoints, making it hard to consider complicated visual styles. 
In contrast, we make use of a texture representation and aim to improve the quality of the UV mapping. Our model is trained without the need to access the textures in advance, which allows us to work with different texture inputs, regardless of complexity. 

\vspace{-2mm}
\paragraph{Generalized video generation.} 
Early methods modelled the entire video clip as a single latent representation~\cite{vondrick2016generating, saito2017temporal}.  Follow-up work MoCoGAN \cite{tulyakov2018mocogan} used a disentangled representation, separating appearance and motion. However, the model is not conditional, so it
cannot generate videos conditioned on target appearances or motions, for example. 
Our method separates appearance and motion by design and allows for easy control and modification of either factor.  We specify appearance via a (fixed) texture map, while  motions are represented by UV coordinates over time.

End-to-end video re-targeting works RecycleGAN\cite{bansal2018recycle} and Vid2Vid\cite{wang2018video} generate videos with content and motion from separate source videos. These methods train target-specific models, in that a new network is trained for each target video. In contrast, re-targeting in our case involves only a simple and efficient look-up.

\section{Preliminaries}
\label{sec:analysis}
\subsection{Notation \& definitions}
An image $I_t \in \mathbb{R}^{s_x \times s_y\times3}$ for frame $t$ in a sequence stores RGB information 
at a location $\mathbf{x} \in \mathbb{R}^{s_x \times s_y}$. The appearance of a person in $I_t$ can also be represented in a texture $T_t \in \mathbb{R}^{t_x \times t_y \times 3}$ with locations $\mathbf{u}$. The image $I_t $ and texture $T_t$ are related via the the UV coordinates $P_t \in \mathbb{R} ^{s_x \times s_y}$, where 
\begin{equation}\label{eq:UVdef}
P_t(\mathbf{x})=\mathbf{u}, \quad \text{s.t.} \quad I_t(\mathbf{x}) = T_t(\mathbf{u}).
\end{equation}

To ensure differentiability, we treat $I_t$, $P_t$ and $T_t$ as continuous functions in space via a suitable interpolation operator; we use bi-linear interpolation in our work. In practice, the three fields are represented as time sequences over $t$. 

The corresponding texture $T_t$ for an image $I_t$ can be generated by warping $I_t$ with function $\mathcal{W}$ via the warping grid $\omega_{T}(P_t)$; conversely, the image content can also be recovered as $I'_t$ from the texture $T_t$ and UV coordinates $P_t$ with warping grid  $\omega_{I}$ from $T_t$ to $I_t$ (see \myreffig{fig:I2T2I}): 
\vspace{-2mm}
\begin{equation}\label{eq:I2T}
\vspace{-2mm}
    T_t = \mathcal{W}(I_t,\;\omega_{T}(P_t)) \quad \text{and} \quad I'_t = \mathcal{W}(T_t,\; \omega_{I}(P_t)).
\end{equation}
\noindent Note the warping function $\mathcal{W}(I,\omega)$,  for every location $\mathbf{x}$ in $I$, returns a bi-linear interpolation of $I$ at location $\omega(\mathbf{x})$.

In our work, we refer to the UV outputs from
DensePose~\cite{alp2018densepose} or unwrapped from the 3D mesh of models like SMPL~\cite{kocabas2020vibe} as \textit{raw} UV coordinates, denoted by $P^r_t$ for frame $t$. Raw UV coordinates are typically restricted by the human body silhouette. As such, loose clothing parts are cut off (see the missing skirt parts in \myreffig{fig:teaser}a and 
\myreffig{fig:I2T2I}a. Additionally, the raw UV $P^r_t$ is not one-to-one. Multiple pixels $\mathbf{x}$ of $I_t$ may be mapped to the same $\mathbf{u}$ in $T_t$, leading to  
a loss of information in $T_t$. 
These two shortcomings may result in extreme and undesirable differences between the original $I_t$ and the reconstructed $I'_t$ (see example in \myreffig{fig:I2T2I}a. 
For a sequence of images over time, the differences are further compounded. As 
$P^r_t$ can only be estimated frame-wise, resulting textures $T_t$ tend to lack correspondence over time.

\subsection{Problem formulation}
Given the non-idealities of $P^r_t$, we aim to develop a system that can output a sequence of refined UV coordinates $P^g_t$ leading to faithful reconstructions $I'_t = I_t$. Additionally, we aim for an independent and lightweight appearance representation in the form of a single texture $T_o$, which is constant over time.

We start by defining a model $G$ parameterized by $\theta$ to estimate refined UV coordinates $P^g_t$ from raw UV $P^r_t$: 
\vspace{-1mm}
\begin{equation}\label{eq:generator}
    P^g_t = G(P^r_t; \theta).
\vspace{-2mm}
\end{equation}
\noindent For $I'_t$ to be of high quality and for $P^g_t$ to be temporally stable, we consider appearance and temporal loss functions
\vspace{-4mm}
\begin{equation}
\begin{aligned}
    \mathcal{L}_{\text{app}} & = \sum_{t=0}^{N}(||I'_t-I_t||^2) 
    =  \sum_{t=0}^{N}(||\mathcal{W}(T_t, \omega_{I}(P^g_t))-I_t||^2),\\
   \mathcal{L}_{\text{temp}} & = \sum_{t=0}^{N}(||T_t -T_{o}||^2) 
   = \sum_{t=0}^{N}( ||\mathcal{W}(I_t, \omega_{T}(P^g_t))-T_{o}||^2),\\
\end{aligned}
 \label{eq:constraints}
\end{equation}
where $N$ represents the sequence length and $T_{o}$ a constant texture. Minimizing $||I'_t-I_t||^2$ 
leads to improvements of $I'_t$. Minimizing $||T_t-T_{o}||^2$ encourages a constant texture, which in turn largely alleviates inconsistent correspondences over time.

\section{Method}
One could learn $\theta$ of model $G$ if raw UV $(P^{r}_t)$ were paired ground truth UV coordinates fulfilling the constraints in \myrefeq{eq:constraints}. Such ground truth data does not exist in practice, so we are forced to consider indirect approaches.  Naively applying an unsupervised or self-supervised training is ill-conditioned and error-prone, due to the strong non-linearities in mappings between $I_t$, $T_t$, and $P_t$.  As such, we propose an approach to combine both supervised and unsupervised learning.  

We start with a data pre-processing step (Sections \ref{sec:extension} to \ref{sec:temporal_relocation}) that gradually refines $P^r_t$ to establish ``ground-truth''.  It is worth noting that we handle the two parts of \myrefeq{eq:constraints} separately due to the strong non-linearity and large distance between $P_t^r$ and $P_t^g$. After an initial training of $G$ with the pre-processed data, we then incorporate unsupervised losses from the image space (Section \ref{sec:training}) to train a final model $G$ that jointly improves spatial and temporal quality. The trained model $G$ generates full-silhouette UV coordinates for different poses.

\you{Since appearance or RGB information is encoded only in the texture $T_o$, which is used for the loss and preprocessing of the data but not a part of the network inputs,} the resulting UV coordinate sequence $P^g_t$ can be directly used for video generation with any given texture. Subsequently, generating a new output sequence with changed colours or patterns is highly efficient.

\subsection{UV extension}
\label{sec:extension}
Raw UV inputs omit important details (see example in \myreffig{fig:I2T2I}a.
First, we aim to achieve full silhouette coverage for $P^r_t$. 
To better understand the relationship between $I_t$, 
$P^r_t$ and $T_t$, we visualize UV mapping results 
for a synthetic grid texture $\mathcal{W}(T_{grid}, \omega_{I}(P^r_t))$ in \myreffig{fig:I2T2I}b.
Here, $T_{grid}$ contains an evenly distributed grid quadrants, which remain well-preserved, suggesting that the UV mapping with $P^r_t$ retains a piece-wise
regular surface manifold, albeit with different scaling factors.

The grid structure suggests that neighbouring points in $I_t$ remain neighbours in $T_t$, and additional entries can be added to the raw UV coordinates $P^r_t$ via extrapolation from neighbouring points.  It is worth pointing out that for traditional UV generation, cutting the object surface and minimizing surface distortion are two challenging steps~\cite{poranne2017autocuts}. The raw UV coordinates provide an initial {unwrapping} of the body, hence we focus on solving the latter challenge of minimizing distortions when extrapolating content in the UV coordinates. 

In this paper, we extend the UV coordinates through energy minimization, employing a \textit{virtual mass-spring} system. Mass-spring systems are commonly used in \you{the simulation of clothing \cite{yang2013cloth,jiang2017anisotropic}. Additionally, \cite{liu2013fast} and \cite{theil2011surface} have shown that the potential energy of a mass-spring system is minimized at the equilibrium state.} 
Due to space limitations, we defer the full exposition to the Supplementary. In our formulation, springs naturally encode the area preservation constraints among new extrapolated points and their neighbouring points in a small region of the texture map. The spring forces drive the new extrapolated points to new positions until the system finds an equilibrium state with reduced distortion. 

\begin{figure}
	\begin{center}
		\includegraphics[width=0.99\linewidth]{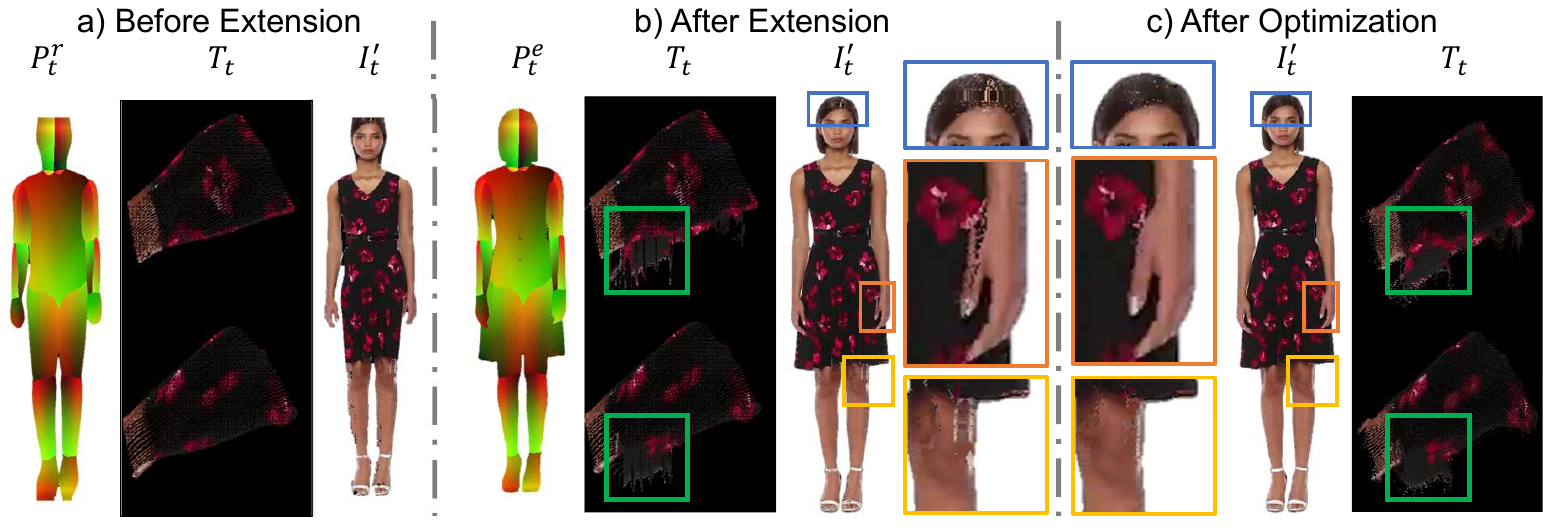}
	\end{center}
	\vspace{-5mm}
	\caption{a) Raw UV coordinates, b) with application of UV extension and c) optimization. The UV extension 
	allows missing parts such as the dress to be mapped into the correct parts of $T_t$, while UV optimization makes $I'_t$ closer to $I_t$.
	}
	\label{fig:P_extrapolation}
	\vspace{-3mm}
\end{figure}

We denote the UV coordinates after the extension with $P^e_t$.  An example result is shown in \myreffig{fig:P_extrapolation}b. We can see that the missing parts from the raw UV coordinates computed via DensePose are recovered successfully, such as the side of the dress.

\subsection{UV optimization}
\label{sec:opt}
After UV extension, artifacts in $I'_{t}$ may remain (See \myreffig{fig:P_extrapolation}b. One cause of these artifacts is duplicate UV coordinates in $P^r_t$, as it is not constrained to be a one-to-one mapping, especially for direct methods such as DensePose. To further improve $P_t$, we directly minimize $\Lapp (P_t)$ via gradient descent, initializing $P_t$ with the extended UV map $P^{e}_t$. The gradient $\frac{\partial \Lapp(P_t)}{\partial P_t}$ can be estimated via the intermediate warping grids $\omega_{T}(P^r_t)$, and $\omega_{I}(P^r_t)$. Details are provided in the Supplementary.

Following common practice in non-linear settings, we add a gradient and Laplacian regularizer to encourage smooth solutions~\cite{belkin2005manifold} and minimize $\Lapp + L_r$, where %, \ie  
\vspace{-2mm}
\begin{equation}
\begin{aligned}
L_{r} = \alpha_1(||\nabla P_t||^2_F) + \alpha_2\sum_{i,j=0,1}||H_{ij}(P_t)||^2_F,
\end{aligned}
\label{eq:smoothing}
\vspace{-2mm}
\end{equation}
$H$ is the Hessian and $||\cdot||_{F}$ denotes the Frobenius norm. It is visible in \myreffig{fig:P_extrapolation}c that most of the artifacts in $I'_t$ have been removed by the optimization procedure, 
and the image content is significantly closer to the reference. \you{We denote the optimized UVs with $P^o_t$}.

\begin{figure*}
\vspace{-2mm}
	\begin{center}
		\includegraphics[width=0.8\linewidth]{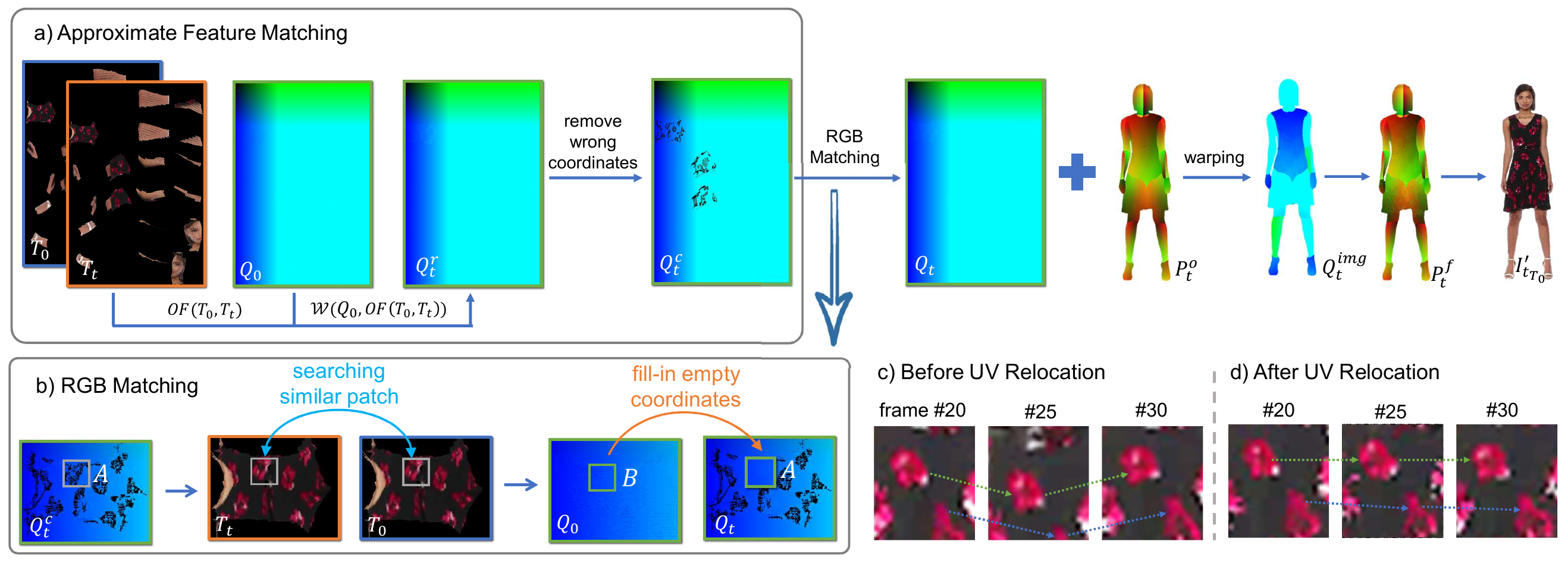}
	\end{center}
	\vspace{-5mm}
	\caption{Overview and results of temporal UV generation. a) Approximate feature matching is achieved via the optical flow (OF) from $T_o$ to $T_t$. b) RGB matching is applied to correct the coordinates resulting from errors in OF. Images in c) are generated with $P^o_t$ and $T_o$, i.e., $I'_{t_{T_o}} = \mathcal{W}(T_o, \omega_{I}(P^o_t))$. Images in d) are similarly generated with $P^f_t$. Green and blue arrows are shown here to track the two patterns in the images. After the temporal relocation step, results are more temporally coherent.  
}
\vspace{-4mm}
	\label{fig:IUV_OF}
\end{figure*}
\subsection{UV temporal relocation}
\label{sec:temporal_relocation}

Minimizing $\Ltemp$ in \myrefeq{eq:constraints} will improve the temporal stability of the texture maps. To do so, we find point correspondences $Q_t(\mathbf{u})$ between $T_t$ and $T_{o}$ so that $T_t(\mathbf{u}) = T_{o}(Q_t(\mathbf{u}))$.  The correspondences allow new UV coordinates $P^f_t$ to map $I_t$ back to the constant $T_{o}$ instead of $T_t$. For simplicity, we assign as the constant $T_o$ the texture from frame 0 of a sequence, \ie $T_o = T_0$. 

We initialize the point correspondences 
with optical flow from $T_o$ to $T_t$, 
\ie $OF(T_o,T_t)$, as shown in \myreffig{fig:IUV_OF}a. An approximate correspondence between $T_t$ and $T_{o}$ can be written as $Q^r_t(\mathbf{u}) =\mathcal{W}(Q_0(\mathbf{u}),OF(T_o,T_t))$. 
In theory, the reconstruction $T'_t$ %$T_t$ 
can then be reconstructed from $T_o$ and $Q^r_t(\mathbf{u})$ via a \emph{lookup} step, \ie 
$T'_t(\mathbf{u}) = T_o(Q^r_t(\mathbf{u}))$.

Note that errors in $OF(T_o,T_t)$ makes  $Q^r_t(\mathbf{u})$ only an approximate correspondence, and there are still differences between the reconstructed $T'_t$ and the true $T_t$.  To correct these errors, we remove the coordinates in $Q^r_t(\mathbf{u})$ where the texture content does not match, \ie $T'_t(\mathbf{u}) \neq T_t(\mathbf{u})$.  We then fill them in with regions from $T_o$ to obtain the final $Q_t(\mathbf{u})$. The filling is based on a simple similarity measure of the RGB values. We defer the details to the Supplementary.

Comparisons of results before and after the temporal relocation step are shown in \myreffig{fig:IUV_OF}c-d. The images $I'_{t_{T_o}}$ recovered from $T_o$ are more temporally coherent after the relocation step.

\subsection{Temporal UV model training}
\label{sec:training}
So far, we have improved the spatial and temporal quality of the raw UV $P^r_t$ separately.  We now consider the two objectives jointly in a spatio-temporal manner and apply an adversarial training for $G$ from~\myrefeq{eq:generator}. The learned $G$ can then generate complete UV coordinates $P^g_t$ at test time given raw UV coordinates $P^r_t$. Below, we define several unsupervised loss terms in both the UV and RGB image space to guide the training and produce high-quality outputs.  
\vspace{-3mm}
\paragraph{Spatial loss (UV space).} 
Recall that $P^f_t$ is now the UV coordinates that relate image $I_t$ to the constant texture $T_o$ based on the UV relocation step in~\myrefsec{sec:temporal_relocation}.  We make use of a supervised $L_2$ loss
$L_{2} = \left \| G(P^r_t) -  P^f_{t} \right \|^2_F$
and adversarial loss via a discriminator $D_{s}$:
\vspace{-2mm}
\begin{equation}
\begin{aligned}
    L^{uv}_{s} &= - log(D_{s}(G(P^r_t))),\\
    L_{D_{s}} &= - log D_{s}(P^f_{t}) - log(1 - D_{s}(G(P^r_t))).
\end{aligned}
\vspace{-3mm}
\end{equation}

\vspace{-3mm}
\paragraph{Temporal stability loss (UV space).} %To improve the temporal stability of the generated UVs, we first 
We consider a smoothing loss between neighbouring frames $t-1$ and $t+1$:
\vspace{-4mm}
\begin{equation}
\begin{aligned}
    L_{smo} = &\left \| G(P^r_{t-1}) - G(P^r_{t})\right \|^2_F + \left \| G(P^r_{t}) - G(P^r_{t+1})\right \|^2_F \\
          &+ \left \| G(P^r_{t-1}) - 2 \times G(P^r_{t}) + G(P^r_{t+1})\right \|^2_F,
\end{aligned}
\end{equation}
\begin{figure}[t]
\vspace{-3mm}
	\begin{center}
		\includegraphics[width=0.85\linewidth]{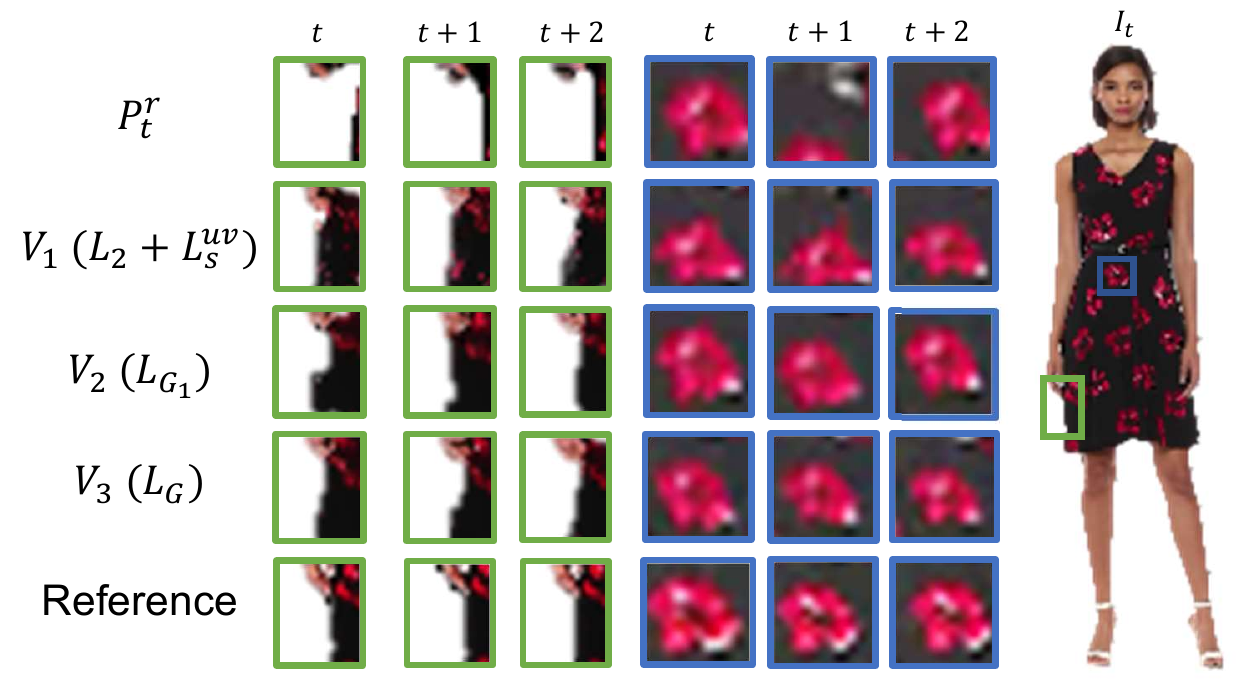}
	\end{center}
	\vspace{-5mm}
	\caption{Comparisons of three successive frames, $(I'_{{t-1}_{T_o}},I'_{t_{T_o}},I'_{{t+1}_{T_o}})$, among results of $P^r_t$, $V_1$, $V_2$, and $V_3$.
	We show examples of the same region in the image to illustrate the temporal coherence of the generated videos. We can see that $V_1$ is more coherent than $P^r_t$ because of the temporal relocation when preparing the training data. $V_2$ and $V_3$ show further improvements due to the temporal stability losses in UV and image spaces. 
	}
	\vspace{-4mm}
	\label{fig:ablation_study}
\end{figure}
and add an unsupervised adversarial loss via a second discriminator network $D_{t}$:
\begin{equation}
\begin{aligned}
    L^{uv}_{t} = &- log(D_{t}(G(P^r_{t-1}),G(P^r_{t}),G(P^r_{t+1}))),\\
    L_{D_{t}} = &- log(D_{t}(P^f_{t-1}),f(P^f_{t-1}),f(f(P^f_{t-1}))) \\
    &- log(1 - D_{t}(G(P^r_{t-1}),G(P^r_{t}),G(P^r_{t+1})) ,
\end{aligned}\label{eq:temploss}
\end{equation}
where $f$ are randomized geometric transformations (e.g., translation, rotation or scaling). Note that ground truth over time is not available in our setting. We synthesize ground truth by randomly choosing a transformation $f$ and applying it to $P^f_{t-1}$. This yields a reference for time $t$; applying the transformation again at $t\!+\!1$ yields an additional reference to form a synthetic triplet.  The triplets serve as ground truth for the adversarial training of \myrefeq{eq:temploss} and guide the generation of smooth UV coordinates over time.

\vspace{-3mm}
\paragraph{Spatial loss (image space).} 
With the mapping pipeline from UV coordinates to images, an image-based L2 loss is applied at training time:
\begin{equation}
\!\!\!\!L^{\text{img}}_{s}\!=\! \left \| I_{g_t}\!-\! I_t \right \|^2_F, \; \text{where} \; I_{g_t}\!=\!\mathcal{W}(T_o, \omega_{I}(G(P^r_{t}))).
\vspace{-3mm}
\end{equation}

\vspace{-3mm}
\paragraph{Temporal stability loss (image space).} Similar to the UV space, we define a temporal adversarial loss via an additional discriminator $D_{img}$ in image space:
\vspace{-2mm}
\begin{equation}
\begin{aligned}
    L^{img}_{t} = &- log D_{img}(I_{g_{t-1}},I_{g_t},I_{g_{t+1}}),\\
    L_{D_{img}} = &- log D_{img}(I_{t-1},I_{t},I_{t+1}) \\
    &- log(1 - D_{img}(I_{g_{t-1}},I_{g_t},I_{g_{t+1}})).
\end{aligned}
\vspace{-2mm}
\end{equation}
\\
To summarize, the full loss of $G$ is given by
\vspace{-1mm}
\begin{equation}
\begin{aligned}
    L_G =  &\lambda_{2}L_{2} + \lambda_{uv,s}L^{uv}_{s} + \lambda_{smo}L_{smo} \\
    &+ \lambda_{uv,t}L^{uv}_{t} + \lambda_{img,s}L^{img}_{s} + \lambda_{img,t} L^{img}_{t}.
\end{aligned}
\label{eq:train_full_loss}
\end{equation}
In practice, we found it difficult to keep the losses in image space stable at the beginning of the training. Hence, we train $G$ first with the partial loss $L_{G_1}$, where
\begin{equation}
\begin{aligned}
 L_{G_1} =  \lambda_{2}L_{2} + \lambda_{uv,s}L^{uv}_{s} + \lambda_{smo}L_{smo} + \lambda_{uv,t}L^{uv}_{t},
\end{aligned}
\end{equation}
for $5\times 10^4$ steps. We freeze $G$, and only train $D_{img}$ for $5\times 10^4$ steps to ensure that  $D_{img}$ is commensurate with $G$.  We then train all networks jointly with the full loss $L_G$ for another $10\times 10^4$ steps.
\huiqi{Generator $G$ is built with ResNet architecture, 
using 30 (for DensePose $P^r_t$) or 20 (for SMPL $P^r_t$) residual blocks.
All of our discriminators $D_s$, $D_t$, and $D_{img}$ follow the same encoder structure using 5 convolutional layers followed by a dense layer. 
We use 120 continuous frames without background from the Fashion dataset~\cite{zablotskaia2019dwnet} as the training data.}  For every step, we randomly crop small regions of size $32\!\times\!32$ from $P^r_t$ to be used as input. The Adam optimizer is applied for training. Other learning details are given in the Supplementary.

\vspace{-3mm}
\paragraph{Model inference.}
After the training, UV coordinates $P^g_t$ with full clothing silhouettes can be generated via $G$. \huiqiRevision{We can achieve pose-guided generation when a sequence of raw target poses is provided.} Since we focus on the UV coordinates and inputs to $G$, which do not include texture information, virtual try-on can also be easily achieved in our pipeline by changing texture maps to which the UV coordinates are applied. Once $P^g_t$ is generated, the image sequence $I'_t$ requires a minimal number of calculations to be produced (essentially, only one texture lookup per output pixel). As we will demonstrate below, this is vastly more efficient than, e.g., evaluating a full CNN.

\begin{table}[]
\begin{center}
\resizebox{0.7\linewidth}{!}{
\begin{tabular}{llllll}
\hline
           & PSNR$\uparrow$  & \begin{tabular}[c]{@{}l@{}}LPIPS$\downarrow$\\ \footnotesize{$\times 10^{-2}$} \end{tabular} & \begin{tabular}[c]{@{}l@{}}tOF$\downarrow$\\ \footnotesize{$\times 10^4$} \end{tabular}   & \begin{tabular}[c]{@{}l@{}}tLP$\downarrow$\\ \footnotesize{$\times 10^{-2}$} \end{tabular}& \begin{tabular}[c]{@{}l@{}}T-diff$\downarrow$\\ \footnotesize{$\times 10^5$} \end{tabular} \\ \hline
\begin{tabular}[c]{@{}l@{}}$P^r_t$ \end{tabular}  & 22.1 & 8.1 & 1.69 & 1.0& 5.42  \\ \hline
$V_1$       & \textbf{23.1} & 7.9 & 1.84 & 1.4 & \textbf{3.93}\\ \hline
$V_2$       & \textbf{23.1} & \textbf{7.6} & 1.70 & \textbf{0.9}  &4.33\\ \hline
$V_3$       & 22.9 & 7.7 & \textbf{1.65} & 1.0  & 4.19\\ \hline
\end{tabular}
}
\end{center}
\vspace{-4mm}
\caption{Quantitative comparisons between $P^r_t$ and our three different versions, $V_1,V_2$, and $V_3$. 
For a fair comparison, the body shapes of $V_1,V_2$, and $V_3$ are cropped to be in line with $P^r_t$. Our method shows significant improvements on both spatial (PSNR and LPIPS) and temporal (tOF, T-diff) evaluation metrics. 
}
\label{tab:black_red_skirt}
\vspace{-5mm}
\end{table}
% \vspace{3mm}
\section{Ablation study}
\label{sec:ablation_study}
This section shows how different parts of \myrefeq{eq:train_full_loss} influence the generated results.
We start with a basic model trained with the losses $L_{2}$ and $L^{uv}_{s}$ and denote this $V_1$. We then add temporal losses in the UV space, $L_{smo}$ and $L^{uv}_{t}$, for training and denote this as $V_2$. The full model trained with $L_G$ is denoted as $V_3$.

\myreffig{fig:ablation_study} shows two qualitative comparisons. All three versions successfully fill in the missing parts of the DensePose UV map and are close to the reference (green patch, skirt edge). To evaluate the temporal coherence, we zoom in on the motion of the flower patterns (blue patch). 
Results from the raw UV coordinates $P^r$ are unsteady since its temporally unstable UV content leads to a misalignment of the texture over time.
$V_1$ has better coherence 
due to the UV relocation  (\myrefsec{sec:temporal_relocation}) applied to the training data. $V_2$ and $V_3$ show progressive improvements 
thanks to the temporal stability losses and the image space losses. 

As quantitative evaluation of the spatial performance, 
we compute peak signal-to-noise ratio (PSNR) and perceptual LPIPS~\cite{zhang2018unreasonable}. For temporal stability, we follow~\cite{chen2017coherent} and estimate the 
differences of warped frames, i.e., $\text{T-diff} = \left \|I_{g_{t}},\mathcal{W}(I_{g_t},v_t)\right \|_1$, \nt{where $v_t$ typically denotes the intra-frame motion computed by optical flow.} In our setting
we use the UV coordinates for $v_t$ instead (details in the Supplementary Material). Additionally, we evaluate with two temporal coherence metrics~\cite{chu2020learning}: $tOF: \left \| OF(I_t,I_{t+1}) - OF(I_{g_t},I_{g_{t+1}}) \right \|_1$ 
and $tLP:  \left \| LPIPS(I_t,I_{t+1}) - LPIPS(I_{g_t},I_{g_{t+1}}) \right \|_1$. 
Except for PSNR, lower values are better for all metrics.

From \myreftab{tab:black_red_skirt}, we see that that $V_1$ has the worst results in terms of tOF and tLP. Its LPIPS is also worse than $V_2$ and $V_3$ because $V_1$ is trained purely with spatial losses in the UV space. Hence, supervision via preprocessed data $P^f_t$ is insufficient. 
Note, however, that $V_1$ shows the best T-diff score, as T-diff mainly relies on the calculation of $v_t$ and is easily ``fooled'' by overly smooth content.
$V_2$ and $V_3$ add temporal constraints and show better temporal behaviour in terms of tOF and tLP.
Compared with $V_2$, $V_3$ exhibits a similar spatial performance though it yields better temporal stability.  This is especially the case if we evaluate without cropping to fit $P^r_t$ (see Supplementary). This also verifies that loss functions from the image space can be successfully applied to guide the training. 

\begin{figure}
\vspace{-2mm}
	\begin{center}
		\includegraphics[width=0.75\linewidth]{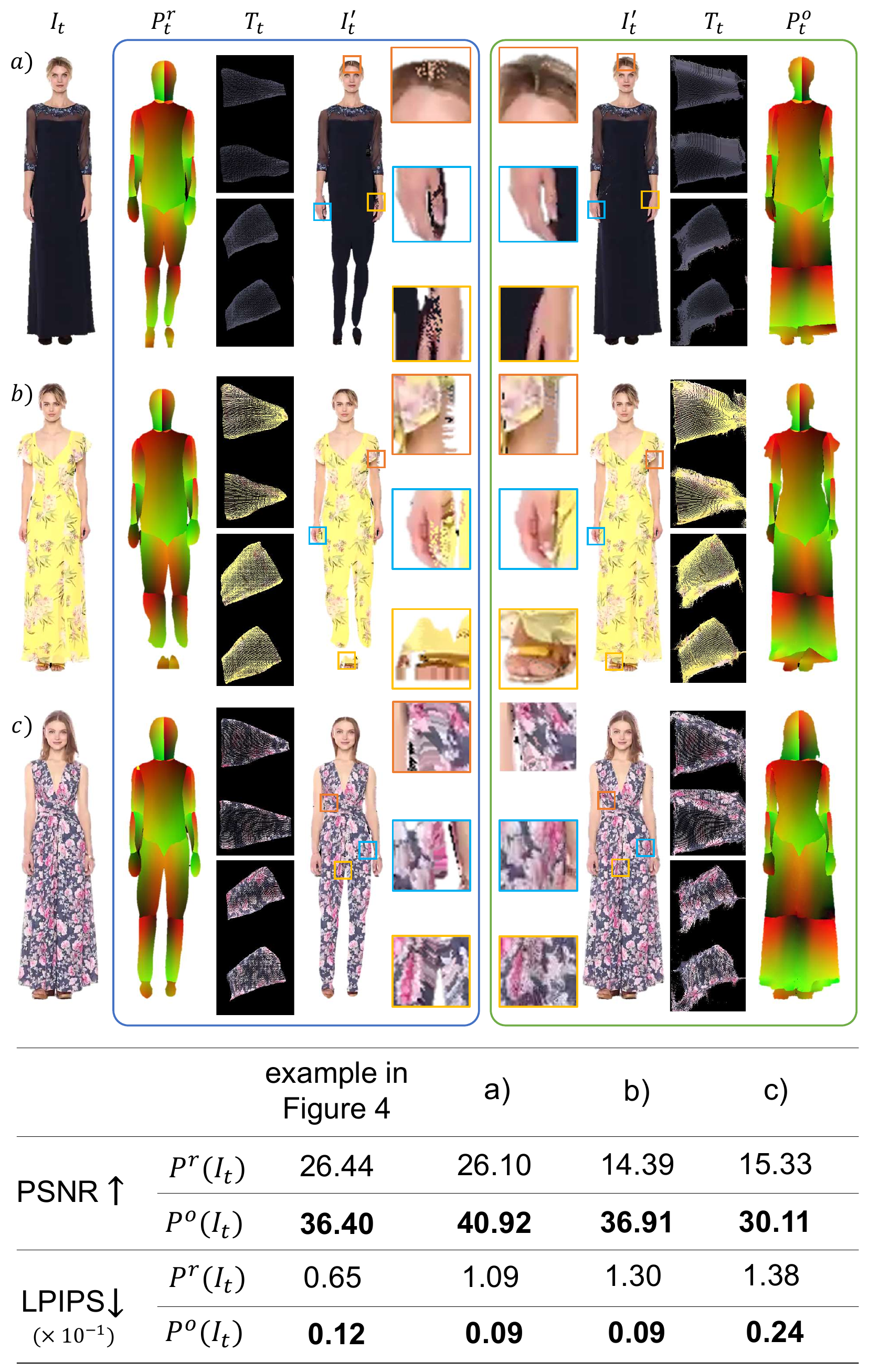}
	\end{center}
	\vspace{-5mm}
	\caption{Comparisons between DensePose UVs $P^r_t$ and optimized UVs $P^o_t$. Here, we only show examples of the skirt part in $T_t$ to clarify the differences. We can see that $P^o_t$ can preserve most of the skirt information in $T_t$, and $I'_t$ of $P^o_t$ are closer to $I_t$ than that of $P^r_t$. The quantitative evaluation also shows that our results after UV optimization (described in \myrefsec{sec:opt}) are closer to the reference.
	}
	\vspace{-4mm}
	\label{fig:more_optimization}
\end{figure}

\begin{figure}[t]
\vspace{-2mm}
	\begin{center}
		\includegraphics[width=0.95\linewidth]{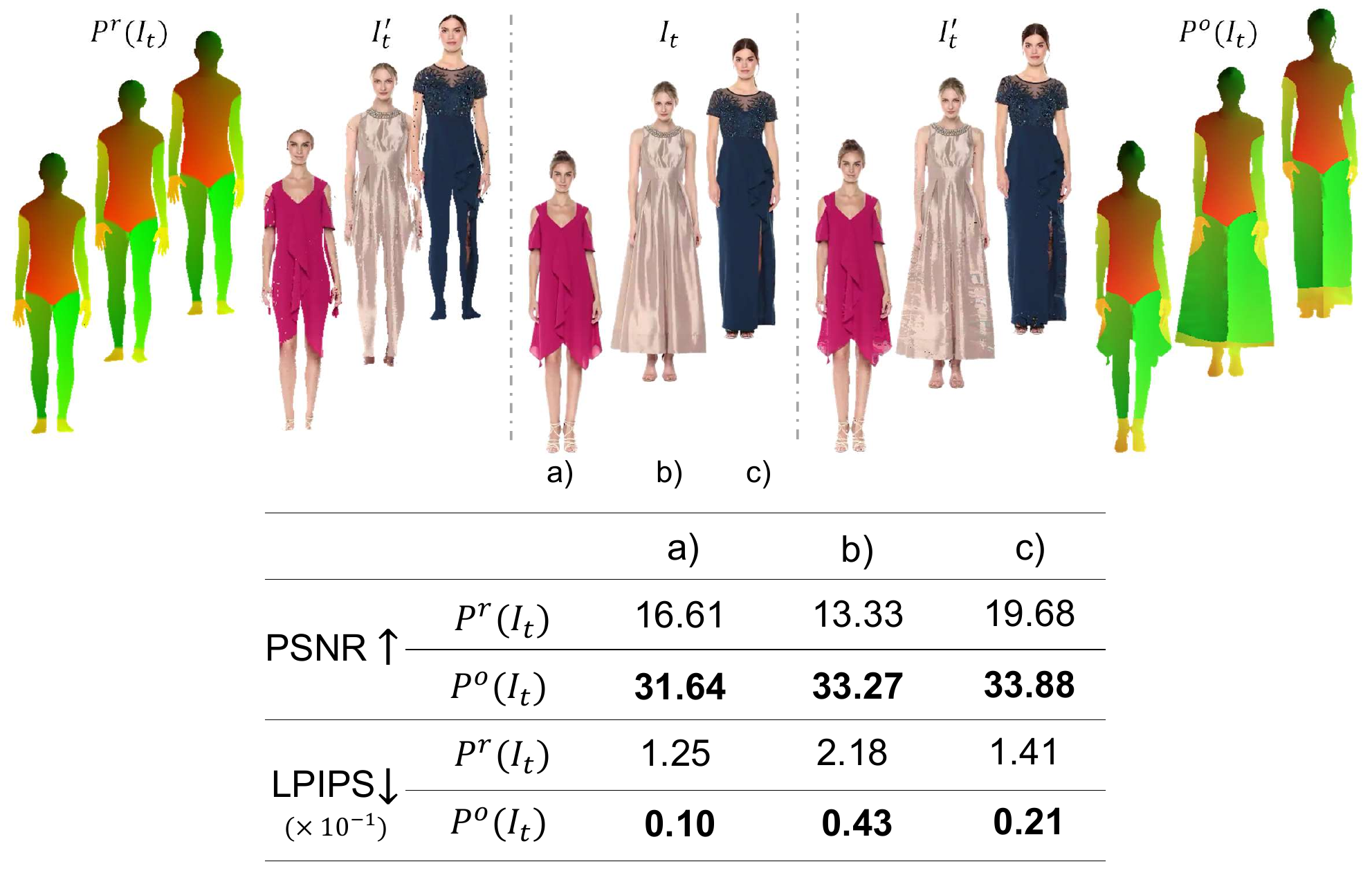}
	\end{center}
	\vspace{-5mm}
	\caption{Comparisons between $P^r_t$ from SMPL and $P^o_t$. We can see that after optimization, $P^o_t$ preserves more of the loose clothing and $I'_t$ closely matches $I_t$. \you{Quantitative evaluations also show that our results are much closer to the reference.}
	}
	\label{fig:more_optimization_smpl}
	\vspace{-4mm}
\end{figure}

\vspace{-4mm}
\paragraph{Optimized UVs ($P^o_t$).}
In addition to \myreffig{fig:P_extrapolation}c \you{in \myrefsec{sec:opt}}, more samples of the optimized UVs $P^o_t$ are shown in \myreffig{fig:more_optimization} and the Supplementary.  
The comparison of PSNR and LPIPS scores in 
\myreffig{fig:more_optimization} 
verifies that our optimization pipeline significantly improves the spatial content. Similar conclusions can be drawn for the UV coordinates derived from SMPL (see \myreffig{fig:more_optimization_smpl}).

\section{Results and evaluation}
\begin{figure*}[ht!]
\vspace{-2mm}
	\begin{center}
		\includegraphics[width=0.78\linewidth]{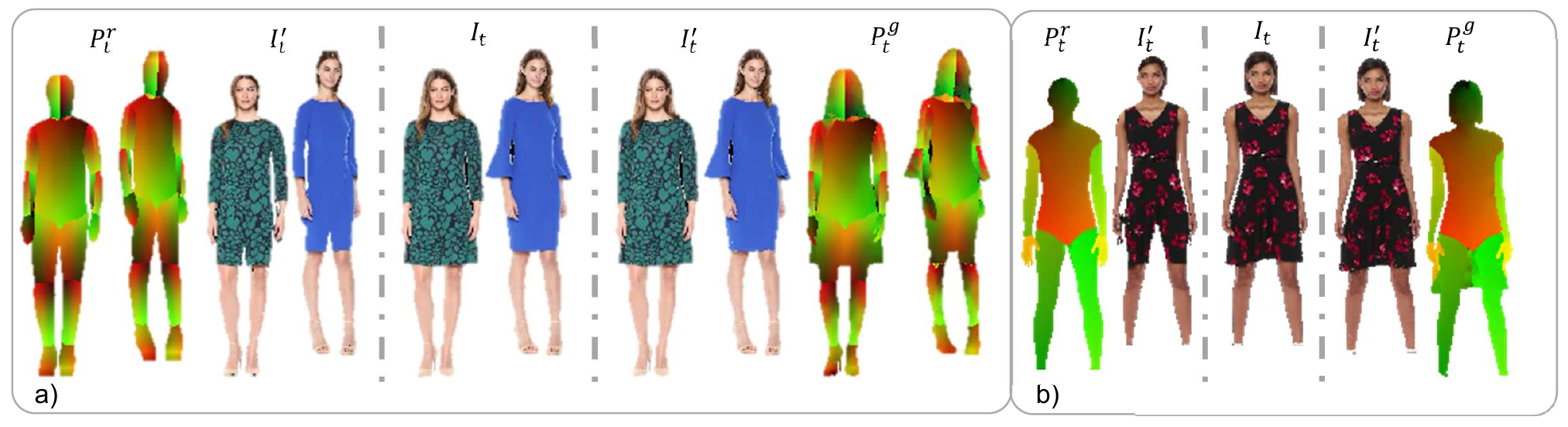}
	\end{center}
	\vspace{-5mm}
	\caption{Comparison between $P^g_t$ and $P^r_t$. $P^g_t$ in (a) and (b) are generated from DensePose and SMPL model, respectively.  Our $I'_t$ are closer to the reference $I_t$, which indicates that $P^g_t$ has better capacity to preserve more information of $I_t$.
	}
	\label{fig:I'_t_comparison}
	\vspace{-3mm}
\end{figure*}

\vspace{-1mm}
\paragraph{Direct comparison of $P^g_t$.}
We provide a direct comparison between raw UVs $P^r_t$ and those generated by our approach $P^g_t$
in \myreffig{fig:I'_t_comparison}. Apart from the body itself, it is visible that our outputs $I'_t$, \you{generated with UVs from both SMPL and DensePose models}, also recover the hair, sleeves, and the skirt. Hence, we have fulfilled the goal of capturing the full appearance of a person, rather than the body silhouette. 

\begin{figure}[h!]
\vspace{-2mm}
	\begin{center}
		\includegraphics[width=0.85\linewidth]{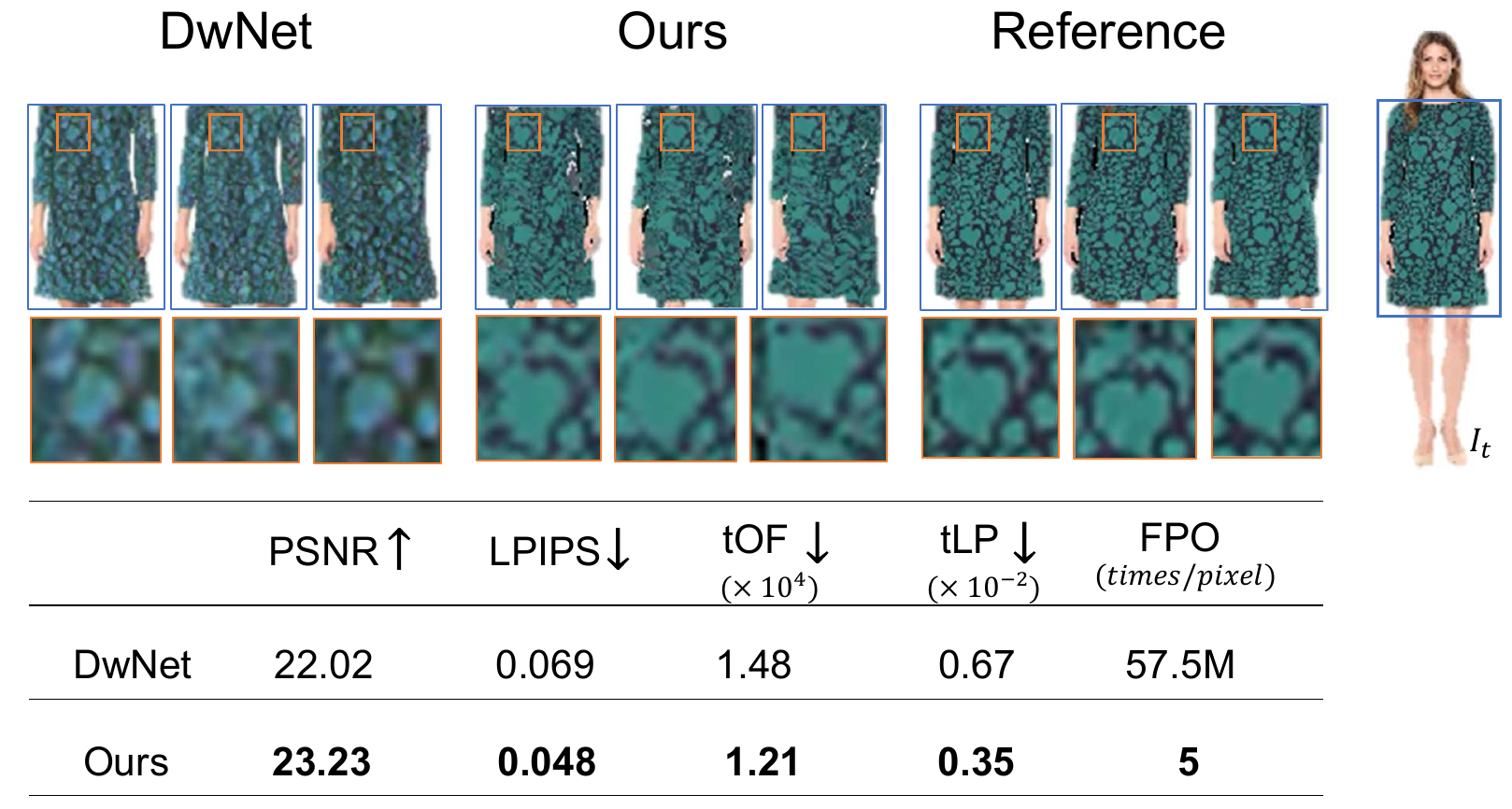}
	\end{center}
	\vspace{-5mm}
	\caption{Comparison with state-of-the-art method DwNet. Our results are closer to the reference, which is also supported by the evaluation metrics below. 
	Additionally, we also compare the number of floating point operations (FPO) for every pixel during video generation. Without rerunning trained models, our method shows a significant reduction of computation.
	}
	\vspace{-4mm}
	\label{fig:dwnet_comparison}
\end{figure}

\vspace{-4mm}
\paragraph{Comparison with state of the art.}
In \myreffig{fig:dwnet_comparison}, we compare with the closest %state-of-the-art algorithm 
method DwNet, which also focuses on video generation from a single image with UV coordinates. 
DwNet smoothes the texture of the clothing as the quality of its output is limited by the accuracy of the warping module. However, our method focuses on UV coordinates, and obtains the appearance information directly from the texture map, so our results are significantly sharper.
Our results are also closer to the reference images, leading to better spatial evaluations like PSNR and LPIPS. For temporal quality, the tOF and tLP values indicate that our results have better temporal stability than DwNet. \youRevision{Consistent conclusions can also be drawn from the user studies illustrated in the Supplementary.}

We note that the DwNet model needs to be rerun once the texture is changed.  In contrast, once the UV coordinates of a sequence have been generated, our method can re-texture a sequence without evaluating any trained models.  Instead, we simply map the updated texture via our UV coordinates; this is a simple lookup that is several orders of magnitude fewer in operations than DwNet.  
Such a low computational load would, \eg, allow for running a virtual try-on pipeline in real-time on otherwise low-performance end-devices. 

\begin{figure}[ht]
\vspace{-2mm}
	\begin{center}
		\includegraphics[width=0.83\linewidth]{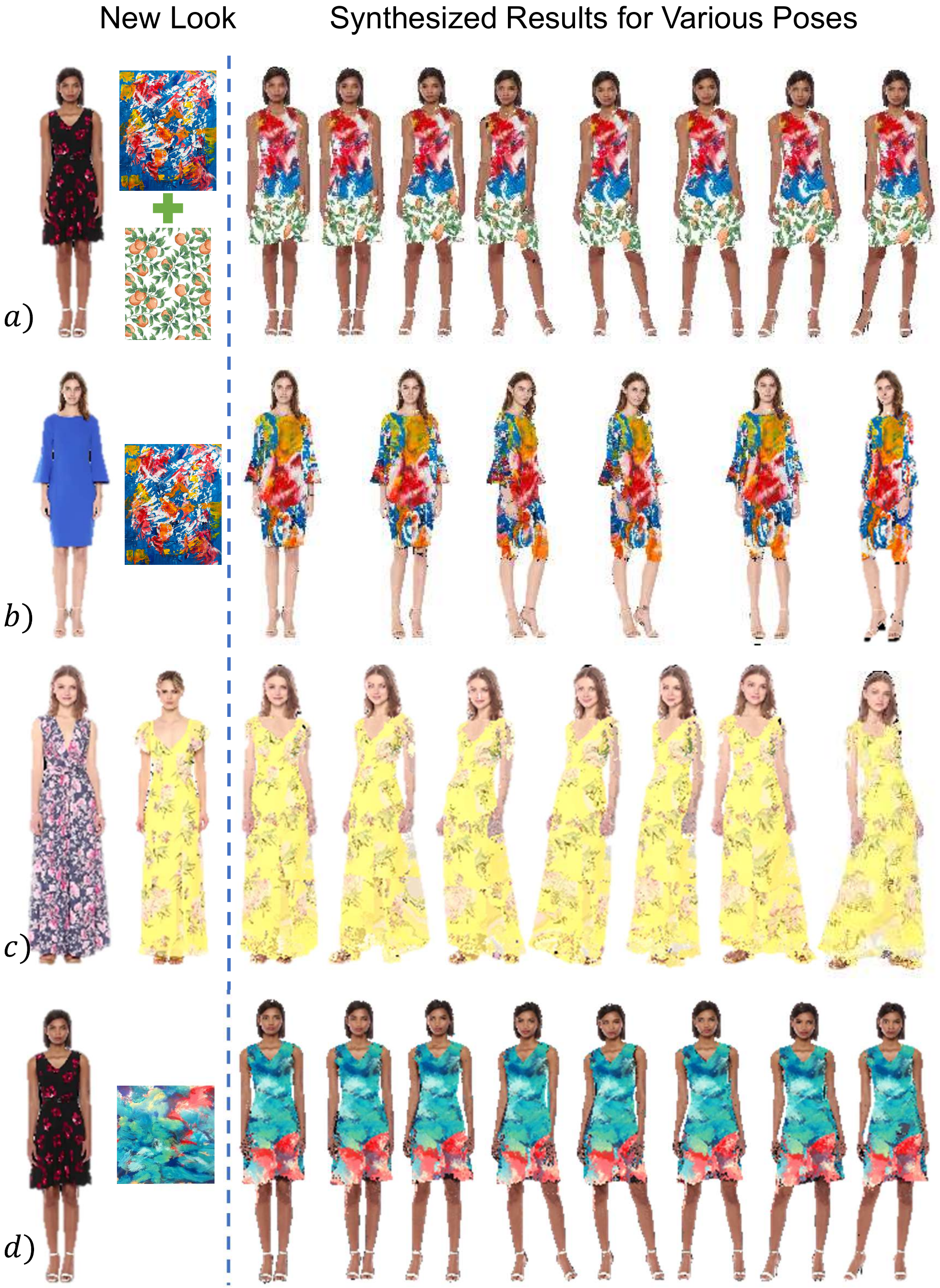}
	\end{center}
	\vspace{-5mm}
	\caption{$P^g_t$ is compatible with different textures to generate a desired target sequence. (a)-(c) are generated using DensePose UV coordinates, while (d) uses SMPL UVs.
	}
	\vspace{-5mm}
	\label{fig:new_texture}
\end{figure}

\vspace{-4mm}
\paragraph{Generated video with different textures.}
Our generation network completely separates the UV representation from the RGB appearance information, which is only encoded in the constant texture $T_o$.  As such, the UV coordinates generated from our model are compatible with any other texture that aligns with the arrangement of the original $T_o$.  This makes it easy to create virtual try-on applications by modifying the texture. In particular, the source clothing can be obtained from any image source, \eg another photo or a texture image.  We show re-textured examples in \myreffig{fig:teaser}, \myreffig{fig:new_texture} and the Supplementary.  Note that as our focus is on capturing clothing, hence we reuse the texture of the human parts (face, hands and legs) from the source videos for these virtual try-on results.

\vspace{-5mm}
\paragraph{Limitations.}
Our method generates an entire video via $P^g_t$ and $T_{o}$.  Currently, we simply choose $T_0$ for $T_{o}$. However, $T_0$ may not have sufficient coverage 
for some situations, \eg the backside of the clothing. 
This could be improved by incorporating additional steps for texture completion~\cite{chibane2020implicit,grigorev2019coordinate}.
\youRevision{Another limitation arises from boundary occlusions. While we aim at coherent point correspondence among different frames, occlusions occurred in the boundary areas make it impossible to find the obscured point at frame $t+1$ for the corresponding point at frame $t$, which brings a noticeable degree of high-frequency noise near the boundaries \you{during fast motions of the body or clothing}. But quantitative metrics and our user study show that our results yield better temporal coherence than state-of-the-art methods. Besides, this problem could benefit from additional image space smoothing over time.}
\section{Conclusion}

We have presented a novel algorithm to generate stable UV coordinates for image sequences that capture the full appearance of a human body, including loose clothing and hair. Central in arriving at this goal are a custom pre-computation pipeline and a spatio-temporal adversarial learning approach.  Our method allows for high-quality video generation and also enables very quick turnaround times for style modifications. Based on the one-time process to generate a UV coordinate sequence, our method allows for the repeated synthesis of output videos via a single underlying texture with vastly reduced computations compared to existing approaches. \youRevision{Currently, we primarily focus on clothing, because the complicated textures and various poses of the body make it a challenging \nt{application}. It provides an appropriate test bed that encapsulates the capabilities of our pipeline. However, our pipeline can be potentially generalized to UVs of other objects, e.g., animals, cars and furniture. This enables us to achieve video generation and texture editing \huiqiRevision{of arbitrary objects} easily in our future work. }
\vspace{-1mm}
\section{Acknowledgement}
This research / project is supported by the Ministry of Education, Singapore, under its MOE Academic Research Fund Tier 2 (STEM RIE2025 MOE-T2EP20220-0015) and the ERC Consolidator Grant \emph{SpaTe} (ERC-2019-COG-863850).

{\small
\bibliographystyle{ieee_fullname}
\bibliography{main}

\begin{thebibliography}{10}\itemsep=-1pt

\bibitem{aberman2019deep}
Kfir Aberman, Mingyi Shi, Jing Liao, Dani Lischinski, Baoquan Chen, and Daniel
  Cohen-Or.
\newblock Deep video-based performance cloning.
\newblock In {\em Computer Graphics Forum}, volume~38, pages 219--233. Wiley
  Online Library, 2019.

\bibitem{alp2018densepose}
R{\i}za Alp~G{\"u}ler, Natalia Neverova, and Iasonas Kokkinos.
\newblock Densepose: Dense human pose estimation in the wild.
\newblock In {\em Proceedings of the IEEE Conference on Computer Vision and
  Pattern Recognition}, pages 7297--7306, 2018.

\bibitem{balakrishnan2018synthesizing}
Guha Balakrishnan, Amy Zhao, Adrian~V Dalca, Fredo Durand, and John Guttag.
\newblock Synthesizing images of humans in unseen poses.
\newblock In {\em Proceedings of the IEEE Conference on Computer Vision and
  Pattern Recognition}, pages 8340--8348, 2018.

\bibitem{bansal2018recycle}
Aayush Bansal, Shugao Ma, Deva Ramanan, and Yaser Sheikh.
\newblock Recycle-gan: Unsupervised video retargeting.
\newblock In {\em Proceedings of the European conference on computer vision
  (ECCV)}, pages 119--135, 2018.

\bibitem{belkin2005manifold}
Misha Belkin, Partha Niyogi, and Vikas Sindhwani.
\newblock On manifold regularization.
\newblock In {\em International Workshop on Artificial Intelligence and
  Statistics}, pages 17--24. PMLR, 2005.

\bibitem{bogo2016keep}
Federica Bogo, Angjoo Kanazawa, Christoph Lassner, Peter Gehler, Javier Romero,
  and Michael~J Black.
\newblock Keep it smpl: Automatic estimation of 3d human pose and shape from a
  single image.
\newblock In {\em European conference on computer vision}, pages 561--578.
  Springer, 2016.

\bibitem{chan2019everybody}
Caroline Chan, Shiry Ginosar, Tinghui Zhou, and Alexei~A Efros.
\newblock Everybody dance now.
\newblock In {\em Proceedings of the IEEE International Conference on Computer
  Vision}, pages 5933--5942, 2019.

\bibitem{chen2017coherent}
Dongdong Chen, Jing Liao, Lu Yuan, Nenghai Yu, and Gang Hua.
\newblock Coherent online video style transfer.
\newblock In {\em Proceedings of the IEEE International Conference on Computer
  Vision}, pages 1105--1114, 2017.

\bibitem{cheng2019multi}
Kun Cheng, Hao-Zhi Huang, Chun Yuan, Lingyiqing Zhou, and Wei Liu.
\newblock Multi-frame content integration with a spatio-temporal attention
  mechanism for person video motion transfer.
\newblock {\em arXiv preprint arXiv:1908.04013}, 2019.

\bibitem{chibane2020implicit}
Julian Chibane and Gerard Pons-Moll.
\newblock Implicit feature networks for texture completion from partial 3d
  data.
\newblock In {\em European Conference on Computer Vision}, pages 717--725.
  Springer, 2020.

\bibitem{chu2020learning}
Mengyu Chu, You Xie, Jonas Mayer, Laura Leal-Taix{\'e}, and Nils Thuerey.
\newblock Learning temporal coherence via self-supervision for gan-based video
  generation.
\newblock {\em ACM Transactions on Graphics (TOG)}, 39(4):75--1, 2020.

\bibitem{deng2018uv}
Jiankang Deng, Shiyang Cheng, Niannan Xue, Yuxiang Zhou, and Stefanos
  Zafeiriou.
\newblock Uv-gan: Adversarial facial uv map completion for pose-invariant face
  recognition.
\newblock In {\em Proceedings of the IEEE conference on computer vision and
  pattern recognition}, pages 7093--7102, 2018.

\bibitem{dong2019fw}
Haoye Dong, Xiaodan Liang, Xiaohui Shen, Bowen Wu, Bing-Cheng Chen, and Jian
  Yin.
\newblock Fw-gan: Flow-navigated warping gan for video virtual try-on.
\newblock In {\em Proceedings of the IEEE International Conference on Computer
  Vision}, pages 1161--1170, 2019.

\bibitem{farras2021rgb}
Albert~Rial Farras, Sergio~Escalera Guerrero, and Meysam Madadi.
\newblock Rgb to 3d garment reconstruction using uv map representations.
\newblock 2021.

\bibitem{gecer2021ostec}
Baris Gecer, Jiankang Deng, and Stefanos Zafeiriou.
\newblock Ostec: One-shot texture completion.
\newblock In {\em Proceedings of the IEEE/CVF Conference on Computer Vision and
  Pattern Recognition}, pages 7628--7638, 2021.

\bibitem{grigorev2018coordinate}
Artur Grigorev, Artem Sevastopolsky, Alexander Vakhitov, and Victor Lempitsky.
\newblock Coordinate-based texture inpainting for pose-guided image generation.
\newblock {\em arXiv preprint arXiv:1811.11459}, 2018.

\bibitem{grigorev2019coordinate}
Artur Grigorev, Artem Sevastopolsky, Alexander Vakhitov, and Victor Lempitsky.
\newblock Coordinate-based texture inpainting for pose-guided human image
  generation.
\newblock In {\em Proceedings of the IEEE/CVF Conference on Computer Vision and
  Pattern Recognition}, pages 12135--12144, 2019.

\bibitem{jiang2017anisotropic}
Chenfanfu Jiang, Theodore Gast, and Joseph Teran.
\newblock Anisotropic elastoplasticity for cloth, knit and hair frictional
  contact.
\newblock {\em ACM Transactions on Graphics (TOG)}, 36(4):1--14, 2017.

\bibitem{kocabas2020vibe}
Muhammed Kocabas, Nikos Athanasiou, and Michael~J Black.
\newblock Vibe: Video inference for human body pose and shape estimation.
\newblock In {\em Proceedings of the IEEE/CVF Conference on Computer Vision and
  Pattern Recognition}, pages 5253--5263, 2020.

\bibitem{liu2019neural}
Lingjie Liu, Weipeng Xu, Michael Zollhoefer, Hyeongwoo Kim, Florian Bernard,
  Marc Habermann, Wenping Wang, and Christian Theobalt.
\newblock Neural rendering and reenactment of human actor videos.
\newblock {\em ACM Transactions on Graphics (TOG)}, 38(5):1--14, 2019.

\bibitem{liu2013fast}
Tiantian Liu, Adam~W Bargteil, James~F O'Brien, and Ladislav Kavan.
\newblock Fast simulation of mass-spring systems.
\newblock {\em ACM Transactions on Graphics (TOG)}, 32(6):1--7, 2013.

\bibitem{liu2019liquid}
Wen Liu, Zhixin Piao, Jie Min, Wenhan Luo, Lin Ma, and Shenghua Gao.
\newblock Liquid warping gan: A unified framework for human motion imitation,
  appearance transfer and novel view synthesis.
\newblock In {\em Proceedings of the IEEE International Conference on Computer
  Vision}, pages 5904--5913, 2019.

\bibitem{loper2015smpl}
Matthew Loper, Naureen Mahmood, Javier Romero, Gerard Pons-Moll, and Michael~J
  Black.
\newblock Smpl: A skinned multi-person linear model.
\newblock {\em ACM transactions on graphics (TOG)}, 34(6):1--16, 2015.

\bibitem{ma2017pose}
Liqian Ma, Xu Jia, Qianru Sun, Bernt Schiele, Tinne Tuytelaars, and Luc
  Van~Gool.
\newblock Pose guided person image generation.
\newblock In {\em Advances in neural information processing systems}, pages
  406--416, 2017.

\bibitem{ma2020unselfie}
Liqian Ma, Zhe Lin, Connelly Barnes, Alexei~A Efros, and Jingwan Lu.
\newblock Unselfie: Translating selfies to neutral-pose portraits in the wild.
\newblock In {\em European Conference on Computer Vision}, pages 156--173.
  Springer, 2020.

\bibitem{madadi2018smplr}
Meysam Madadi, Hugo Bertiche, and Sergio Escalera.
\newblock Smplr: Deep smpl reverse for 3d human pose and shape recovery.
\newblock {\em arXiv preprint arXiv:1812.10766}, 2018.

\bibitem{neverova2018dense}
Natalia Neverova, Riza Alp~Guler, and Iasonas Kokkinos.
\newblock Dense pose transfer.
\newblock In {\em Proceedings of the European conference on computer vision
  (ECCV)}, pages 123--138, 2018.

\bibitem{neverova2019slim}
Natalia Neverova, James Thewlis, Riza~Alp Guler, Iasonas Kokkinos, and Andrea
  Vedaldi.
\newblock Slim densepose: Thrifty learning from sparse annotations and motion
  cues.
\newblock In {\em Proceedings of the IEEE/CVF Conference on Computer Vision and
  Pattern Recognition}, pages 10915--10923, 2019.

\bibitem{pavlakos2018learning}
Georgios Pavlakos, Luyang Zhu, Xiaowei Zhou, and Kostas Daniilidis.
\newblock Learning to estimate 3d human pose and shape from a single color
  image.
\newblock In {\em Proceedings of the IEEE Conference on Computer Vision and
  Pattern Recognition}, pages 459--468, 2018.

\bibitem{poranne2017autocuts}
Roi Poranne, Marco Tarini, Sandro Huber, Daniele Panozzo, and Olga
  Sorkine-Hornung.
\newblock Autocuts: simultaneous distortion and cut optimization for uv
  mapping.
\newblock {\em ACM Transactions on Graphics (TOG)}, 36(6):1--11, 2017.

\bibitem{pumarola2018unsupervised}
Albert Pumarola, Antonio Agudo, Alberto Sanfeliu, and Francesc Moreno-Noguer.
\newblock Unsupervised person image synthesis in arbitrary poses.
\newblock In {\em Proceedings of the IEEE Conference on Computer Vision and
  Pattern Recognition}, pages 8620--8628, 2018.

\bibitem{pumarola2019unsupervised}
Albert Pumarola, Vedanuj Goswami, Francisco Vicente, Fernando De~la Torre, and
  Francesc Moreno-Noguer.
\newblock Unsupervised image-to-video clothing transfer.
\newblock In {\em Proceedings of the IEEE/CVF International Conference on
  Computer Vision Workshops}, pages 0--0, 2019.

\bibitem{saito2017temporal}
Masaki Saito, Eiichi Matsumoto, and Shunta Saito.
\newblock Temporal generative adversarial nets with singular value clipping.
\newblock In {\em Proceedings of the IEEE international conference on computer
  vision}, pages 2830--2839, 2017.

\bibitem{siarohin2019animating}
Aliaksandr Siarohin, St{\'e}phane Lathuili{\`e}re, Sergey Tulyakov, Elisa
  Ricci, and Nicu Sebe.
\newblock Animating arbitrary objects via deep motion transfer.
\newblock In {\em Proceedings of the IEEE Conference on Computer Vision and
  Pattern Recognition}, pages 2377--2386, 2019.

\bibitem{siarohin2019first}
Aliaksandr Siarohin, St{\'e}phane Lathuili{\`e}re, Sergey Tulyakov, Elisa
  Ricci, and Nicu Sebe.
\newblock First order motion model for image animation.
\newblock In {\em Advances in Neural Information Processing Systems}, pages
  7137--7147, 2019.

\bibitem{siarohin2018deformable}
Aliaksandr Siarohin, Enver Sangineto, St{\'e}phane Lathuiliere, and Nicu Sebe.
\newblock Deformable gans for pose-based human image generation.
\newblock In {\em Proceedings of the IEEE Conference on Computer Vision and
  Pattern Recognition}, pages 3408--3416, 2018.

\bibitem{tang2020xinggan}
Hao Tang, Song Bai, Li Zhang, Philip~HS Torr, and Nicu Sebe.
\newblock Xinggan for person image generation.
\newblock In {\em European Conference on Computer Vision}, pages 717--734.
  Springer, 2020.

\bibitem{theil2011surface}
Florian Theil.
\newblock Surface energies in a two-dimensional mass-spring model for crystals.
\newblock {\em ESAIM: Mathematical Modelling and Numerical Analysis},
  45(5):873--899, 2011.

\bibitem{tulyakov2018mocogan}
Sergey Tulyakov, Ming-Yu Liu, Xiaodong Yang, and Jan Kautz.
\newblock Mocogan: Decomposing motion and content for video generation.
\newblock In {\em Proceedings of the IEEE conference on computer vision and
  pattern recognition}, pages 1526--1535, 2018.

\bibitem{vondrick2016generating}
Carl Vondrick, Hamed Pirsiavash, and Antonio Torralba.
\newblock Generating videos with scene dynamics.
\newblock In {\em Advances in neural information processing systems}, pages
  613--621, 2016.

\bibitem{wang2018video}
Ting-Chun Wang, Ming-Yu Liu, Jun-Yan Zhu, Guilin Liu, Andrew Tao, Jan Kautz,
  and Bryan Catanzaro.
\newblock Video-to-video synthesis.
\newblock {\em arXiv preprint arXiv:1808.06601}, 2018.

\bibitem{wang2020imaginator}
Yaohui Wang, Piotr Bilinski, Francois Bremond, and Antitza Dantcheva.
\newblock Imaginator: Conditional spatio-temporal gan for video generation.
\newblock In {\em Proceedings of the IEEE/CVF Winter Conference on Applications
  of Computer Vision}, pages 1160--1169, 2020.

\bibitem{yan2021ultrapose}
Haonan Yan, Jiaqi Chen, Xujie Zhang, Shengkai Zhang, Nianhong Jiao, Xiaodan
  Liang, and Tianxiang Zheng.
\newblock Ultrapose: Synthesizing dense pose with 1 billion points by
  human-body decoupling 3d model.
\newblock In {\em Proceedings of the IEEE/CVF International Conference on
  Computer Vision}, pages 10891--10900, 2021.

\bibitem{yang2018pose}
Ceyuan Yang, Zhe Wang, Xinge Zhu, Chen Huang, Jianping Shi, and Dahua Lin.
\newblock Pose guided human video generation.
\newblock In {\em Proceedings of the European Conference on Computer Vision
  (ECCV)}, pages 201--216, 2018.

\bibitem{yang2013cloth}
Jian~Dong Yang and Shu~Yuan Shang.
\newblock Cloth modeling simulation based on mass spring model.
\newblock In {\em Applied Mechanics and Materials}, volume 310, pages 676--683.
  Trans Tech Publ, 2013.

\bibitem{yoon2021pose}
Jae~Shin Yoon, Lingjie Liu, Vladislav Golyanik, Kripasindhu Sarkar, Hyun~Soo
  Park, and Christian Theobalt.
\newblock Pose-guided human animation from a single image in the wild.
\newblock In {\em Proceedings of the IEEE/CVF Conference on Computer Vision and
  Pattern Recognition}, pages 15039--15048, 2021.

\bibitem{zablotskaia2019dwnet}
Polina Zablotskaia, Aliaksandr Siarohin, Bo Zhao, and Leonid Sigal.
\newblock Dwnet: Dense warp-based network for pose-guided human video
  generation.
\newblock {\em arXiv preprint arXiv:1910.09139}, 2019.

\bibitem{zhang2018unreasonable}
Richard Zhang, Phillip Isola, Alexei~A Efros, Eli Shechtman, and Oliver Wang.
\newblock The unreasonable effectiveness of deep features as a perceptual
  metric.
\newblock In {\em Proceedings of the IEEE conference on computer vision and
  pattern recognition}, pages 586--595, 2018.

\bibitem{zhu2020simpose}
Tyler Zhu, Per Karlsson, and Christoph Bregler.
\newblock Simpose: Effectively learning densepose and surface normals of people
  from simulated data.
\newblock In {\em European Conference on Computer Vision}, pages 225--242.
  Springer, 2020.

\end{thebibliography}
}

\newpage
\appendix
\hspace{-4mm}{\Large{\textbf{Supplementary}}}\\

In the following, we will first illustrate more details about UV preprocessing, such as UV extension (\myrefsec{sec:app_extension}), UV optimization (\myrefsec{sec:app_opt}), and UV temporal relocation (\myrefsec{sec:app_temporal_relocation}). Then, additional details about our training will be given in \myrefsec{sec:app_training}. In \myrefsec{sec:app_evaluation} and \myrefsec{sec:app_more_results}, we will further discuss our evaluation and results.

\section{UV extension}
\label{sec:app_extension}
\begin{figure}
	\begin{center}
		\includegraphics[width=0.7\linewidth]{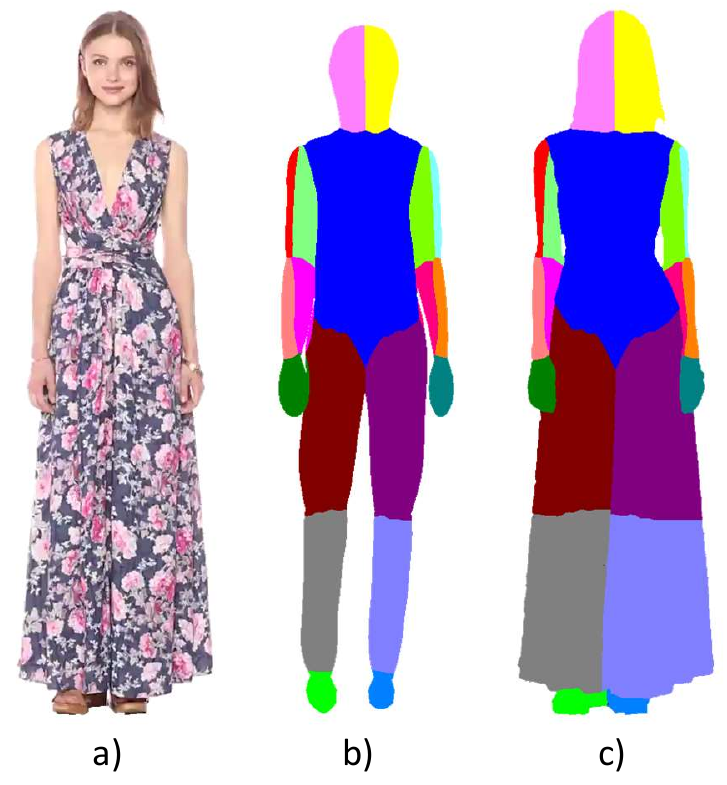}
	\end{center}
	\caption{Example of a) $I_t$ and b) $I$ channel from DensePose $P^r_t$; c) after the labelling, the $I$ channel is extended to the full silhouette.
	}
	\label{fig:labelling}
\end{figure}

In this section, we provide further details about UV extension (\myrefsec{sec:extension} of the main paper). Given an image, we first remove the background to obtain $I_t$. Note that the DensePose model is actually an IUV model, \ie the UV coordinates $P^r_t$ has an additional $I$ channel that encodes the \you{24} individual parts of the body (see \you{\myreffig{fig:labelling}b}).   
Subsequently, we unwrap the surface in a part-by-part manner based on the individual parts before applying UV extrapolation to $P^r_t$. 
We assign the labels of the extended parts \you{manually according to their location}, 
e.g. we label the hair on the left side with the same $I$ values as the left head part. An example of our labelling is shown in \myreffig{fig:labelling}.

After labelling the $I$ values, we linearly extrapolate values for  empty positions with neighbouring points with the same $I$ value inside a small area with size $3 \times 3$. Then we map the new extrapolated point from $I_t$ to $T_t$ and apply a \textit{virtual mass-spring} system in $T_t$ to reduce the surface distortion. \you{We assume that all neighbouring points $O_1,O_2,...,O_n$ inside a region of size $40 \times 40$ are connected with this new extrapolated point $O_0$ via \textit{virtual} springs in the texture. The pushing/pulling forces $\mathbf{f_1},\mathbf{f_2},...,\mathbf{f_n}$ from neighbour points will drive $O_0$ to the direction of $\sum_{i=1}^{n}\mathbf{f_i}$ for every step until $O_0$ arrives at an equilibrium state with a new position, where $\sum_{i=1}^{n}\mathbf{f_n} = \mathbf{0}$.}

From our experience, we found that applying pure pushing forces can generate valid results, since the parts that need to be extended are always located at the outline of the body.
After applying pushing forces, we switch the forces to pulling in order to make the texture more compact. To summarize, we extend UV coordinates in the UV space and then continue with a \textit{virtual mass-spring} system in the texture map to reduce the unwrapped surface distortion. For $P^r_t$ generated from the SMPL model, which only contains UV channels, the UV extension step is performed without labelling the $I$ values, i.e. we directly extend $P^r_t$ to the full silhouette with linear extrapolation before applying the \textit{virtual mass-spring} system.

\begin{figure*}
	\begin{center}
		\includegraphics[width=0.9\linewidth]{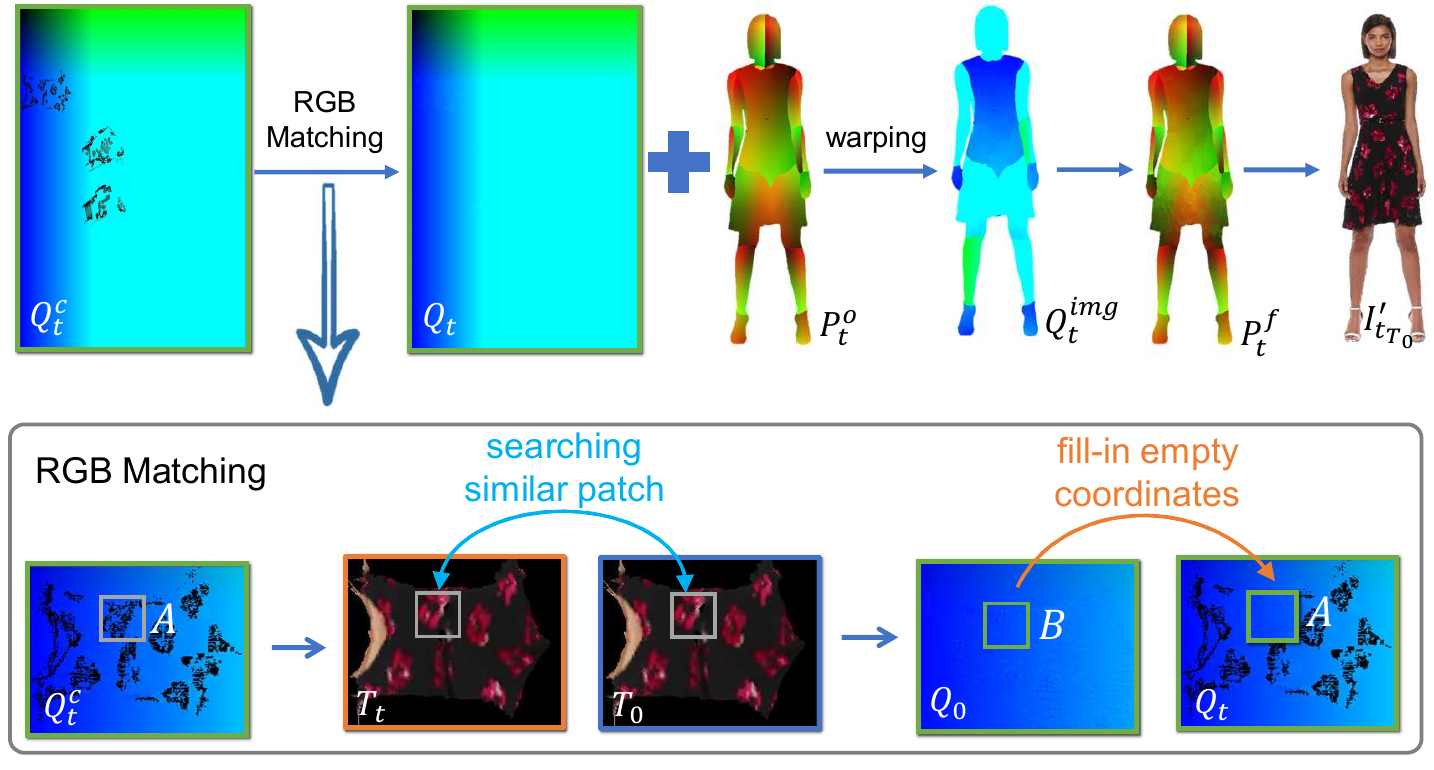}
	\end{center}
	\caption{RGB matching step of UV temporal relocation. Empty positions in $Q^c_t$ are filled via patch matching between $T_0$ and $T_t$.
	}
	\label{fig:app_IUV_OF}
\end{figure*}

\section{UV optimization}
\label{sec:app_opt}
\you{In \myrefsec{sec:opt} of the main paper, we discussed the steps to obtaining $P^o_t = \argmin \left \| I_t - I_t^{'} \right \|_F^2$.}
In this section, we will illustrate how we calculate $\frac{\partial \mathcal{L}_{\text{app}}}{\partial P_t}$ to optimize $P_t$ with the objective function
\begin{equation}
    \mathcal{L}_{\text{app}} = \left \| I_t - I_t^{'} \right \|_F^2,
\end{equation}
from which initially we will have
\begin{equation}
\begin{aligned}
    \frac{\partial \mathcal{L}_{\text{app}}}{\partial P_t} = \frac{\partial \mathcal{L}_{\text{app}}}{\partial I'_t} \times \frac{\partial I'_t}{\partial P_t}.
\end{aligned}
\end{equation}

We compute the gradient $\frac{\partial I'_t}{\partial P_t}$ via the intermediate warping grids $\omega_{T}(P^r_t)$ and $\omega_{I}(P^r_t)$ from UV mappings 
\begin{equation}\label{eq:app_I2T}
    T_t = \mathcal{W}(I_t,\;\omega_{T}(P_t)) \quad \text{and} \quad I'_t = \mathcal{W}(T_t,\; \omega_{I}(P_t)).
\end{equation}
And relationship between $\omega_{I}(P_t)$ and $P_t$ can be written as
\begin{equation}
    \omega_{I}(P_t(\mathbf{x})) = \mathbf{x} - P_t(\mathbf{x}).
\label{eq:uv2coor}
\end{equation}
This gives:
\begin{equation}
\begin{aligned}
\frac{\partial I'_t}{\partial P_t} &= \frac{\partial I'_t}{\partial T_t}  \times \frac{\partial T_t}{\partial P_t}  + \frac{\partial I'_t}{\partial \omega_{I}(P_t)} \times \frac{\partial \omega_{I}(P_t)}{\partial P_t};\\
\frac{\partial T_t}{\partial P_t} &= \frac{\partial T}{\partial  \omega_{T}(P_t)} \times \frac{\partial  \omega_{T}(d(I_t))}{\partial P_t};\\
\frac{\partial  \omega_{T}(P_t)}{\partial P_t} &= \frac{\omega_{T}(P_t)}{\partial \omega_{I}(P_t)} \times \frac{\partial \omega_{I}(P_t)}{\partial P_t}.
\end{aligned}
\end{equation}
In an implementation, we can conveniently obtain $\frac{\partial  \omega_{T}(P_t)}{\partial P_t}$ via
\begin{equation}
\frac{\partial  \omega_{T}(P_t)}{\partial P_t}  = \mathcal{W}(\frac{\partial \omega_{I}(P_t)}{\partial P_t}, \omega_{T}(P_t)),
\end{equation}
and $\frac{\partial \omega_{I}(P_t)}{\partial P_t}$ can be computed via \myrefeq{eq:uv2coor}. This provides $\frac{\partial I'_t}{\partial P_t}$ for optimization and learning steps. 

We apply a gradient descent optimizer with $\alpha_1\!=\!100$ and $\alpha_2\!=\!10$ for regularizer $L_r$. Due to the large distance between $P^e_t$ and $P^o_t$, we use a large learning rate, such as $10.0$, to accelerate the optimization procedure. We found that  promising results can be obtained after ca. 16500 steps. \youRevision{UV optimization takes about 75s/frame, measured for resolution $1200 \times 800$ with a NVIDIA RTX 2080 Ti GPU. All frames can be optimized in parallel.}

\section{UV temporal relocation}
\label{sec:app_temporal_relocation}
In this section, we will introduce how we use RGB matching in \myrefsec{sec:temporal_relocation} of the main paper to fill in the empty areas in $Q^c_t$.
Specifically, we assume that similar, nearby texture patches in $T_0$ and $T_t$ will have the same correspondences in $Q_0$ and $Q_t$. For a missing area $A$ in $Q^c_t$ , as shown in \myreffig{fig:app_IUV_OF}, we locate the region with the same position as $A$ in $T_t$. \you{We record the values of $Q^c_t$ and $T_t$ inside region $A$ with $[Q^c_t]_A$ and $[T_t]_A$, respectively. Then we can find a region $B$ in $T_0$ via}
\begin{equation}
min\ ||[T_t]_A - [T_0]_B||_F^2,
\end{equation}
and $[Q_0]_B$ are used to fill in $[Q^c_t]_A$ to obtain $Q_t$.

Afterwards, $Q_t$ can be mapped to $Q^{img}_t$ in the image space via $P^o_t$. $Q^{img}_t(\mathbf{u})$ is directly our $P^f_t$ if $P^r_t$ is represented with the same coordinates system as $\mathbf{u}$, such as the $P^r_t$ unwrapped from the SMPL model. But for the $P^r_t$ from the DensePose model, which uses a different coordinate system from $\mathbf{u}$, we additionally transform $Q^{img}_t(\mathbf{u})$ into $P^f_t$.

\section{Training details}
\label{sec:app_training}
\begin{figure*}
	\begin{center}
		\includegraphics[width=0.99\linewidth]{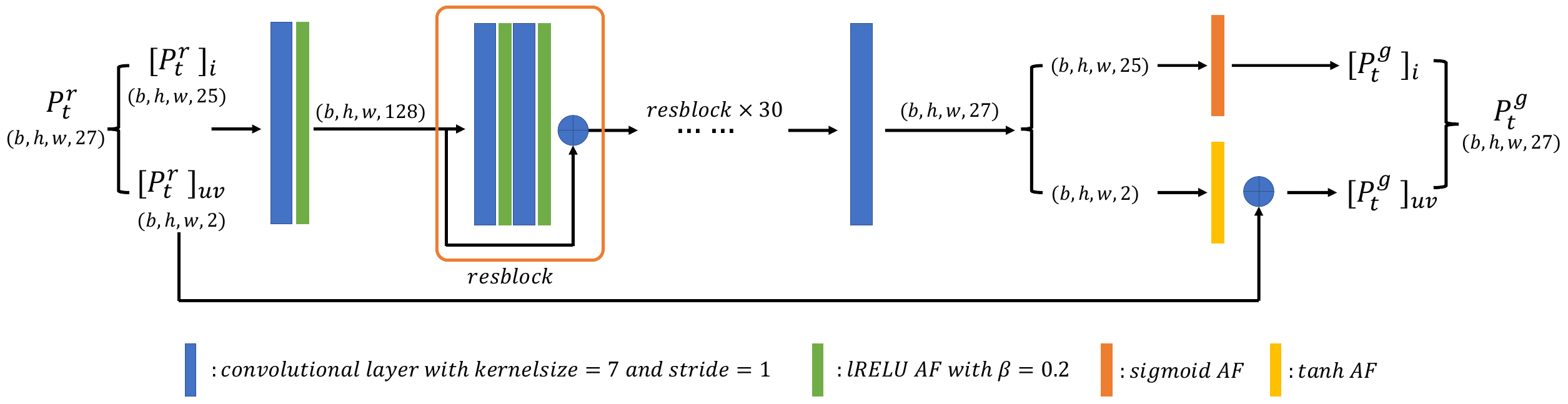}
	\end{center}
	\caption{Generator structure for training with $P^r_t$ from the DensePose model. The corresponding part of $I$ channel will be removed when training with $P^r_t$ generated from the SMPL model.
	}
	\label{fig:generator}
\end{figure*}

\begin{figure}
	\begin{center}
		\includegraphics[width=0.99\linewidth]{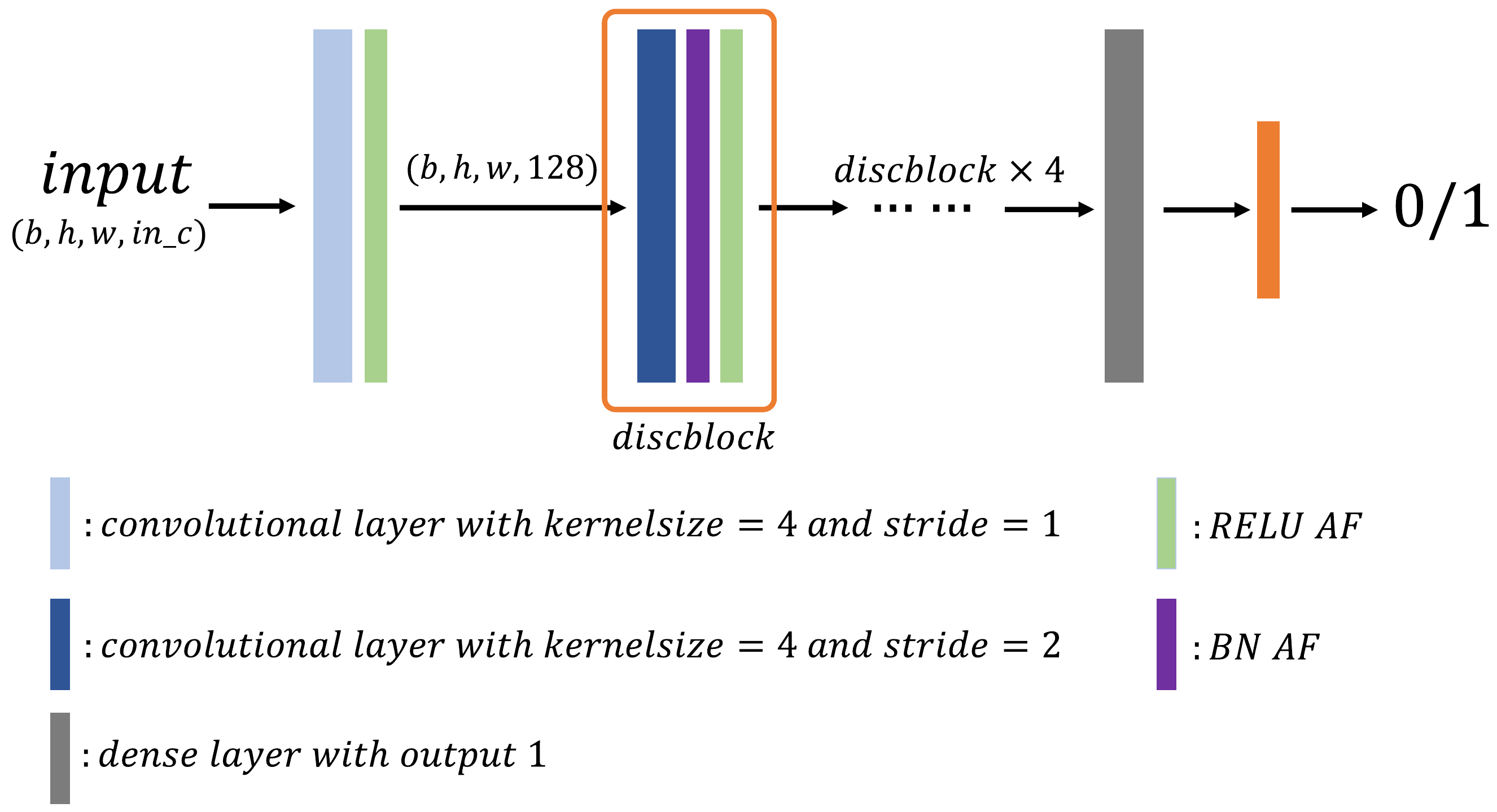}
	\end{center}
	\caption{Architecture of the discriminator networks, such as $D_s$, $D_t$, and $D_{img}$. Input channels $in_c$ for $D_s$, $D_t$, and $D_{img}$ are 2, 6, and 9, respectively.
	}
	\label{fig:discriminator}
\end{figure}

In this section, more details about temporal UV model training (\myrefsec{sec:training} of the main paper) will be illustrated. 
The DensePose model outputs UV coordinates with an extra $I$ channel to classify different body parts. Below,  we use \you{$I$ subscripts to denote the $I$ channel of a UV coordinate, e.g., $G_{I}(P^r_t)$ refers to the $I$ channel of $G(P^r_t)$}.
Then, for $P^r_t$ from the DensePose model, we use an extra cross-entropy loss
\begin{equation}
    L_I =  -\textit{\textbf{e}}_{[P^f_{t}]_I} log G_{I}(P^r_t)
\end{equation}
for $I$ channel constraint, where $-\textit{\textbf{e}}_{[P^f_{t}]_I}$ is a one-hot vector indicating the $[P^f_{t}]_I^{\text{th}}$ $I$ channel with $-\textit{\textbf{e}}_{[P^f_{t}]_{Ij}} = 1$ if $j = [P^f_{t}]_I$. We train $G$ for DensePose $P^r_t$ with the architecture shown in \myreffig{fig:generator}, and all of the discriminators $D_s$, $D_t$, and $D_{img}$ follow the same encoder structure, as shown in \myreffig{fig:discriminator}. 
For UV data without an $I$ channel, e.g., $P^r_t$ generated from SMPL models, our pipeline is still applicable by training without $L_i$ and removing the $I$ channel part in $G$.

We apply gradient clipping for the gradients from $L^{img}_{s}$ and $L^{img}_{t}$ to stabilize the training of $G$. Parameters $\lambda_2$ and $\lambda_{uv,s}$ start from $200$ and $10$, respectively. They are decreased with rate $0.99$ for every 1000 steps. On the other hand, $\lambda_{img,s}$, $\lambda_{smo}$, $\lambda_{uv,t}$, and $\lambda_{img,t}$ start from $0.001$, $0.1$, $1$, and $1$, respectively, but they are gradually increased with rate $1.01$ for every 1000 steps.
\begin{figure}
	\begin{center}
		\includegraphics[width=0.4\linewidth]{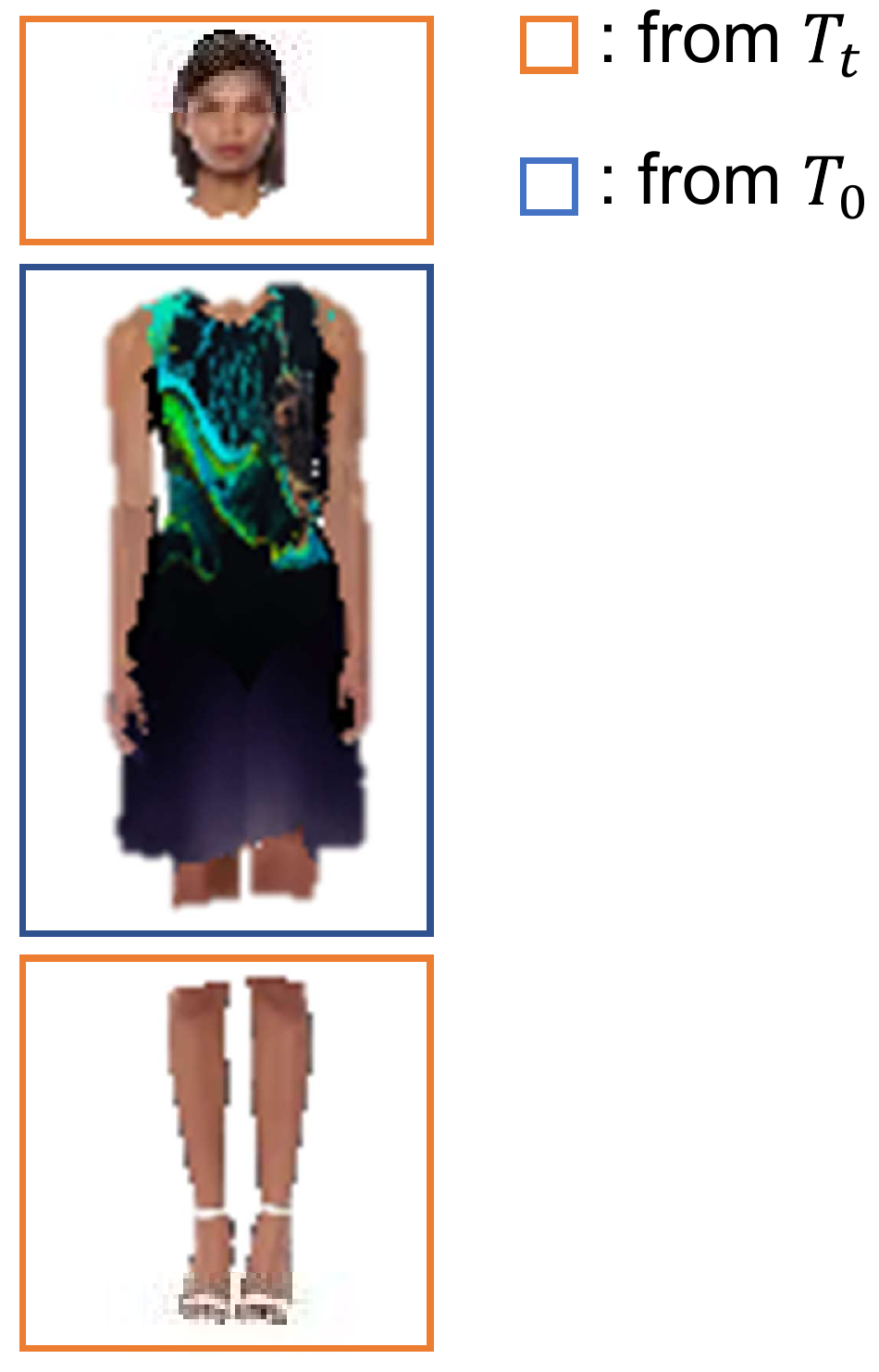}
	\end{center}
	\caption{For parts that do not interact with the clothing, such as head and feet (in the yellow rectangles), we reuse texture of $T_t$ to reconstruct those parts in $I'_t$. For the rest of the parts (blue rectangle), we use the constant texture $T_o$.
	}
	\label{fig:synthesizing}
\end{figure}
\section{Evaluation of results}
\label{sec:app_evaluation}

\begin{figure}
	\begin{center}
		\includegraphics[width=0.99\linewidth]{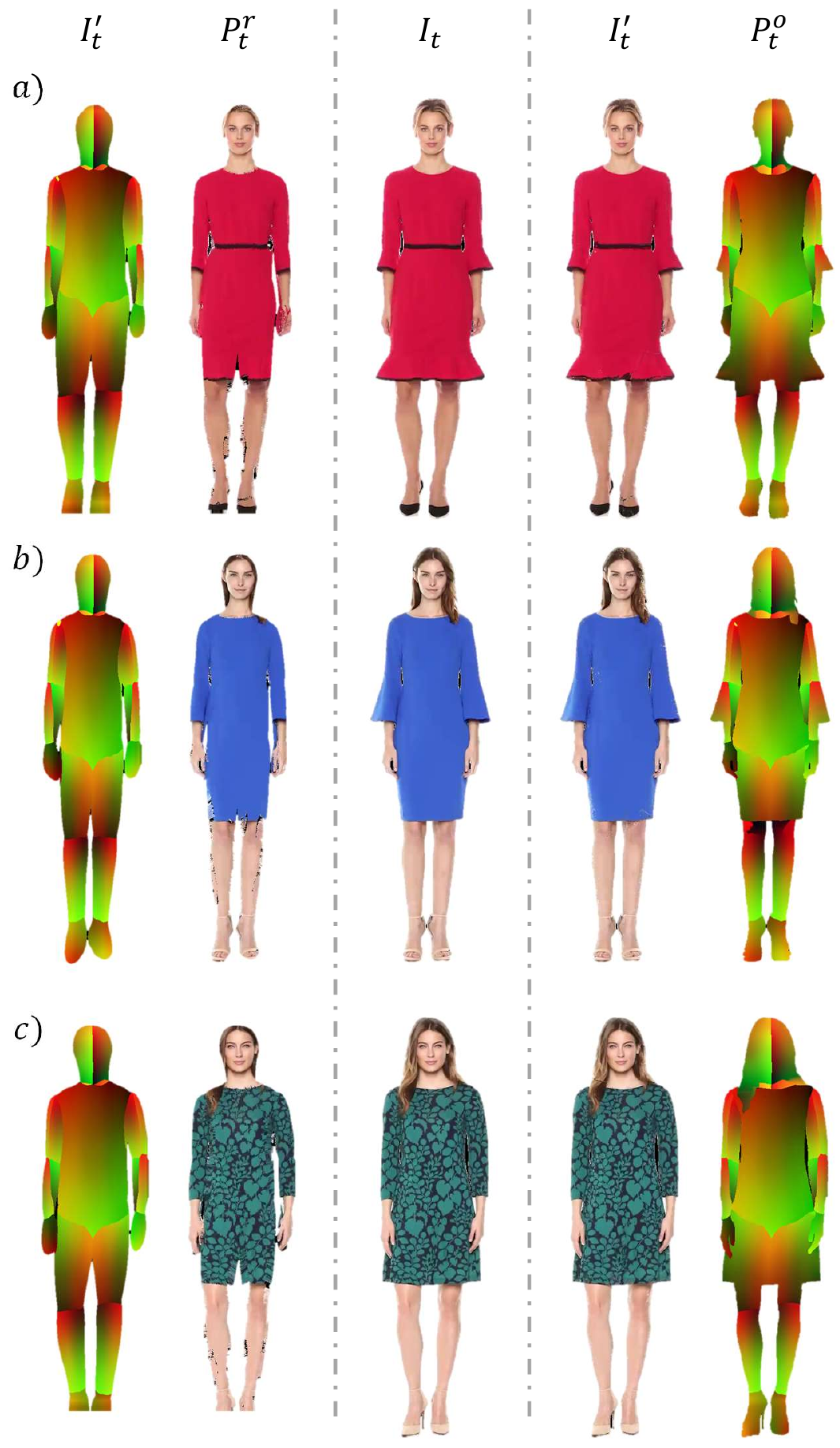}
	\end{center}
	\caption{\you{Additional results comparing raw UV coordinates $P^r_t$ (the fist column) with UV coordinates $P^o_t$ (the fifth column) after our optimization step. Here we also show $I'_t$ for $P^r_t$ (the second column) and $P^o_t$ (the fourth column).} We can see that the results $I'_t$ generated with $P^o_t$ are closer to the reference $I_t$.
	}
	\label{fig:app_more_optimization}
\end{figure}

\begin{figure*}[t!]
	\begin{center}
		\includegraphics[width=0.9\linewidth]{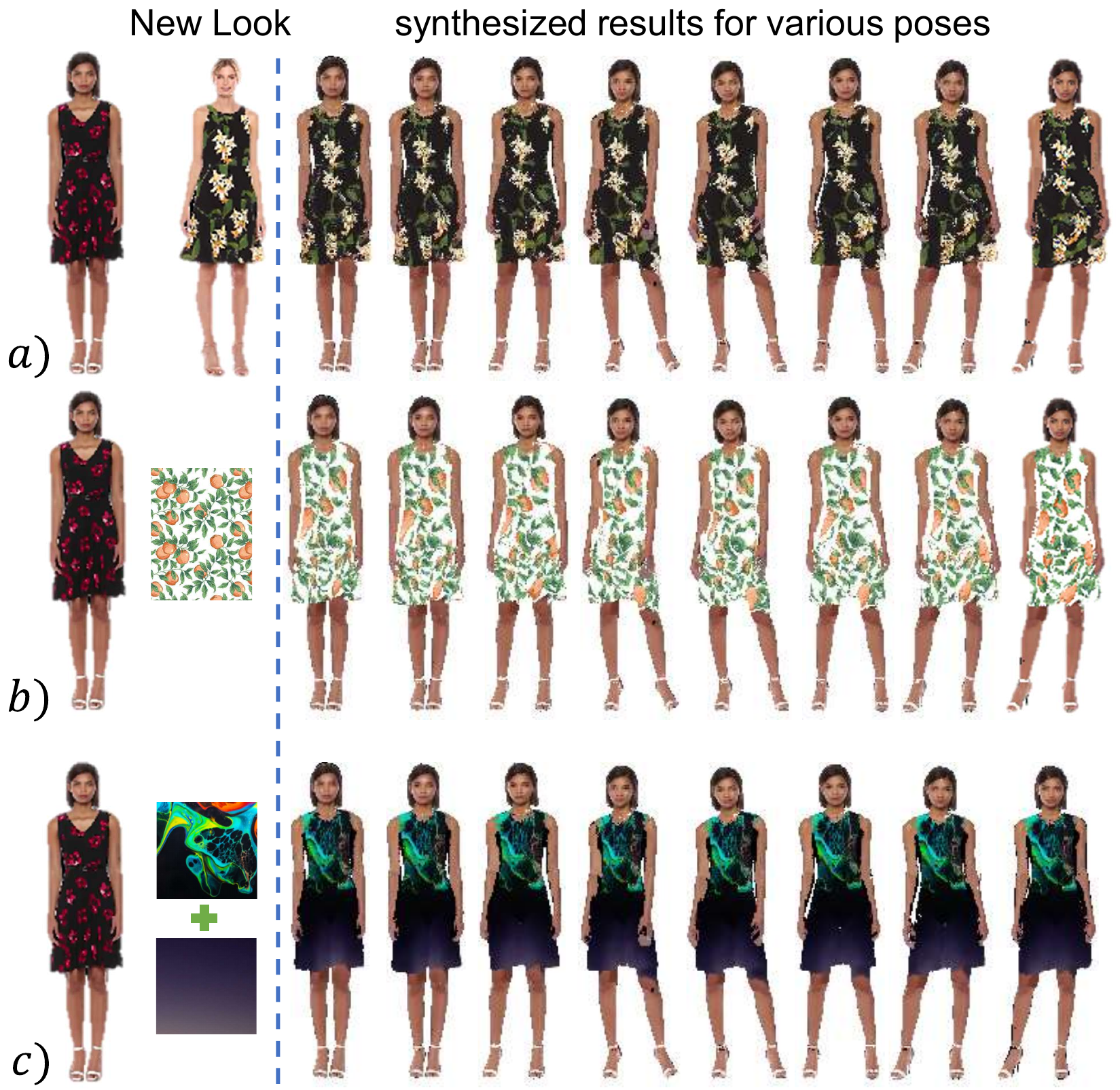}
	\end{center}
	\caption{Additional virtual try-on results. Different textures, regardless of complexity, can be applied as a new look to our source video very efficiently.
	}
	\label{fig:more_try_on}
\end{figure*}

\begin{table}[]
\begin{center}
\resizebox{0.8\linewidth}{!}{
\begin{tabular}{llllll}
\hline
           & PSNR$\uparrow$  & \begin{tabular}[c]{@{}l@{}}LPIPS$\downarrow$\\ \footnotesize{$\times 10^{-2}$} \end{tabular} & \begin{tabular}[c]{@{}l@{}}tOF$\downarrow$\\ \footnotesize{$\times 10^4$} \end{tabular}   & \begin{tabular}[c]{@{}l@{}}tLP$\downarrow$\\ \footnotesize{$\times 10^{-2}$} \end{tabular}& \begin{tabular}[c]{@{}l@{}}T-diff$\downarrow$\\ \footnotesize{$\times 10^5$} \end{tabular} \\ \hline
\begin{tabular}[c]{@{}l@{}}$P^r_t$ \end{tabular}  & 22.1 & 8.1 & 1.69 & \textbf{1.0}& 5.42  \\ \hline
$V_1$       & 23.8 & 7.0 & 1.95 & 1.7 & \textbf{4.33}\\ \hline
$V_2$       & \textbf{23.9} & \textbf{6.7} & 1.76 & 1.3  &4.67\\ \hline
$V_3$       & 23.6 & 6.8 & \textbf{1.68} & 1.2  & 4.55\\ \hline
\end{tabular}
}
\end{center}
\vspace{-4mm}
\caption{Quantitative comparisons between $P^r_t$ and our different versions without cropping to fit =$P^r_t$. Our method show significant improvements for both spatial (PSNR and LPIPS) and temporal (tOF, T-diff) evaluation metrics. Evaluations with full shape lead to further improved PSNR and LPIPS evaluations for our results. 
}
\label{tab:app_black_red_skirt}
\end{table}

In \myrefsec{sec:ablation_study} of the main paper, we follow ~\cite{chen2017coherent} to evaluate temporal coherence of the results and estimate the 
differences of warped frames, i.e., $\text{T-diff} = \left \|I_{g_{t}},\mathcal{W}(I_{g_t},v_t)\right \|_1$. In our setting,
we use the UV coordinates to calculate $v_t$, so that $\text{T-diff}$ will purely be influenced by $P_t$. We first warp all the point coordinates $\mathbf{x}$ in $I_{g_t}$ to the texture space, then we can calculate the displacement of all the points from $I_{g_t}$ to $I_{g_{t+1}}$: 
\begin{equation}
v^{texture}_t = \mathcal{W}(c_{img},\omega_{T}(P^g_{t+1}))- \mathcal{W}(c_{img},\omega_{T}(P^g_t))),
\end{equation}
where $c_{img}(\mathbf{x}) = \mathbf{x}$.
Then $v_t$ can be obtained with:
\begin{equation}
v_t = \mathcal{W}(v^{texture}_t,\omega_{I}(P^g_{t+1})).
\end{equation}

\begin{figure}[H]
	\begin{center}
	\vspace{-6mm}
		\includegraphics[width=0.6\linewidth]{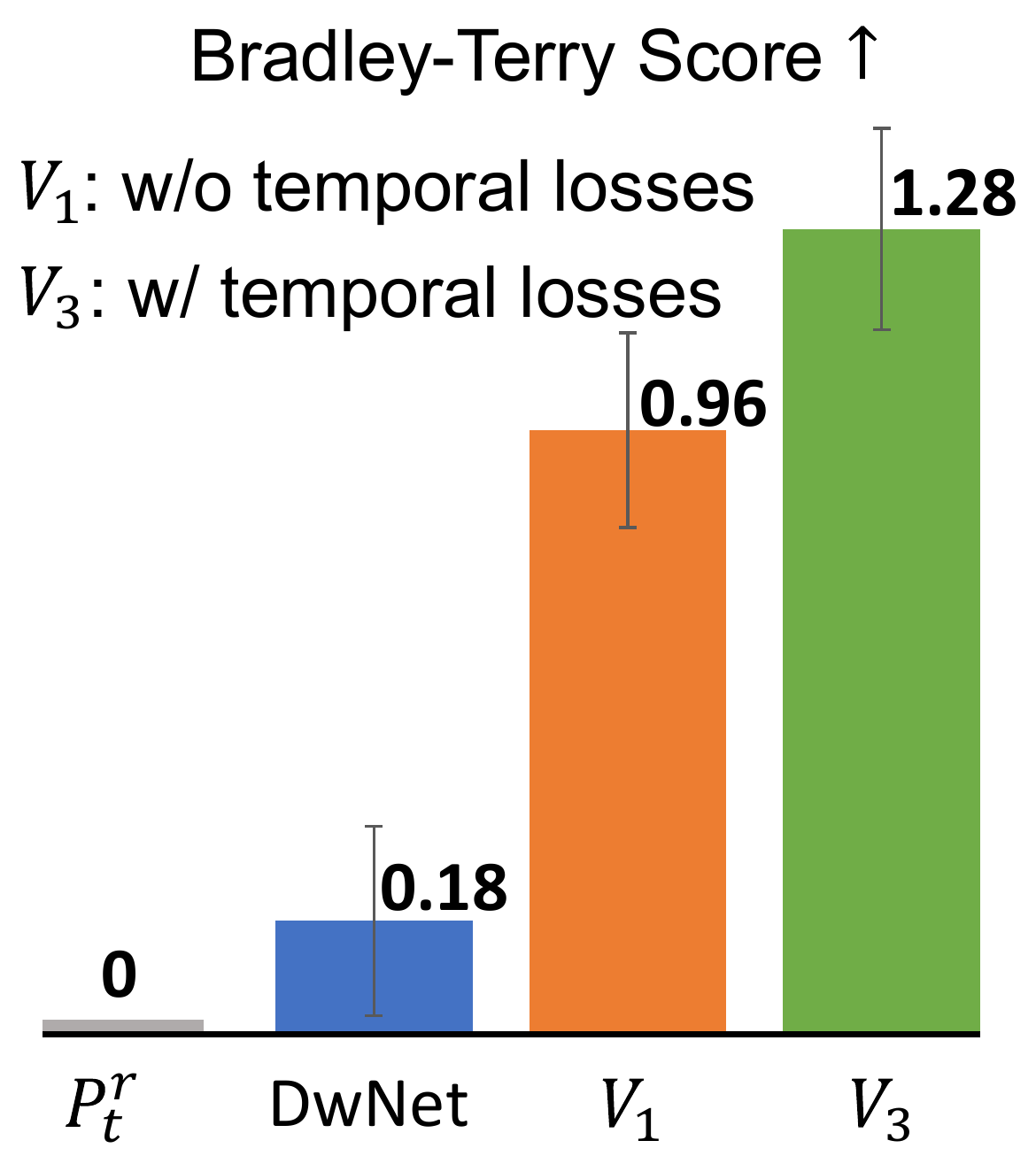}
	\end{center}
	\vspace{-6mm}
	\caption{\youRevision{User study for the red-black dress \nt{case}. Our full version $V_3$ significantly improves over $V_1$ 
	and DwNet.}
	}
	\vspace{-3mm}
	\label{fig:new_result}
\end{figure}
\begin{figure}[H]
	\begin{center}
	\vspace{-5mm}
		\includegraphics[width=0.99\linewidth]{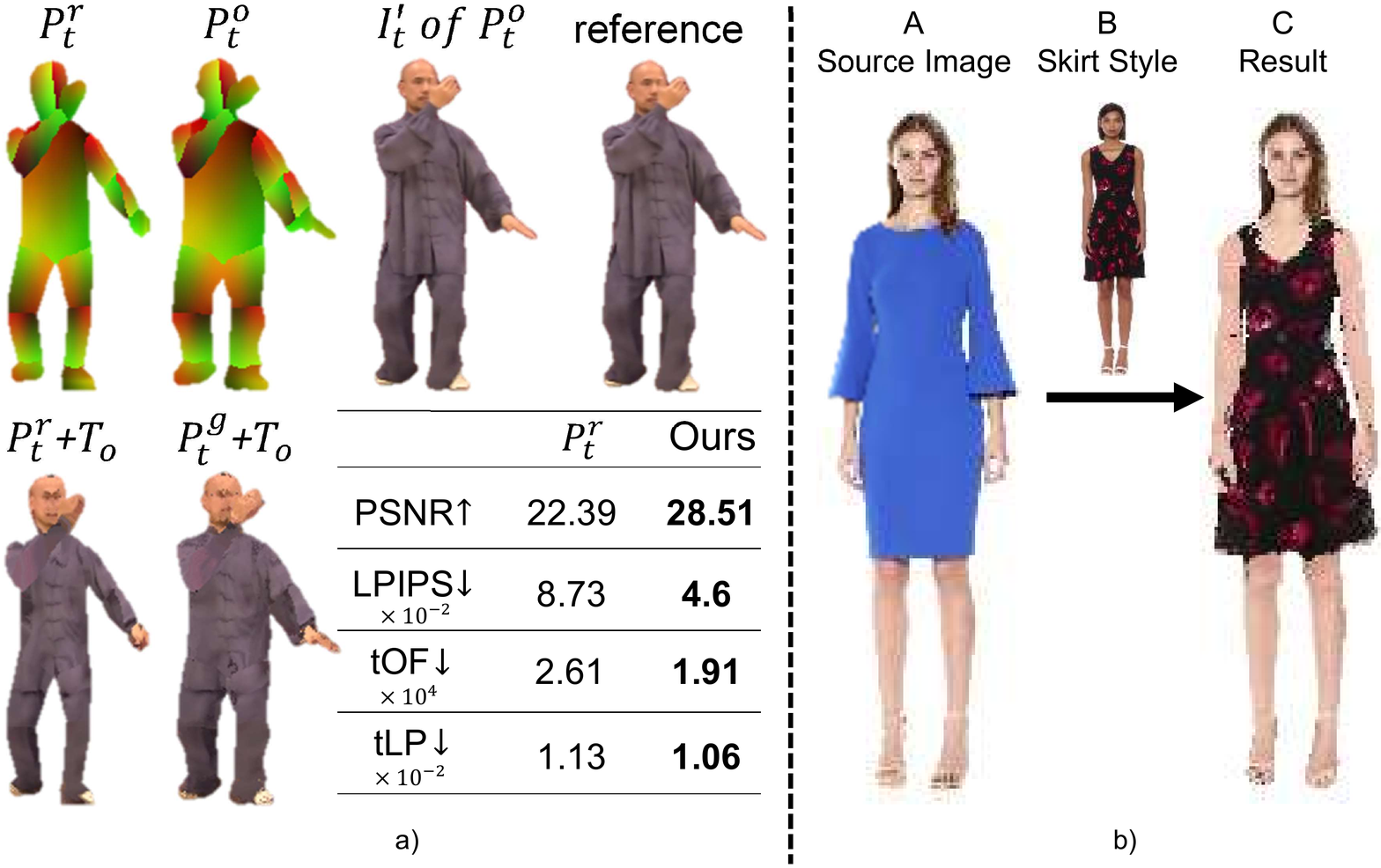}
	\end{center}
	\vspace{-6mm}
	\caption{\youRevision{a) Results of the Tai-Chi dataset. b) \huiqiRevision{Pose-guided generation application. }Our model is generalized to different poses from different videos.}}
	\vspace{-3mm}
	\label{fig:new_transfer}
\end{figure}

In \myreftab{tab:black_red_skirt} of the main paper, we show  quantitative comparisons between $P^r_t$ and our different versions, which are made fair by cropping to fit =$P^r_t$. These results show improvements for both spatial and temporal evaluations. Here, we also show comparisons of those versions without cropping in \myreftab{tab:app_black_red_skirt}. We can see that our versions outperform $P^r_t$ even further in terms of spatial quality. $P^r_t$ performs the best with tLP, as $P^r_t$ cannot generate the extended skirt and hair parts, which significantly decreases the area for evaluation. Here, we also can see that $V_3$ shows similar spatial quality as $V_2$ but an improved temporal coherence. Please refer to the supplementary video to see the improved temporal coherence of the synthesized sequence.

\youRevision{We conducted a user study to evaluate coherence
(see \myreffig{fig:new_result}. Raw DensePose $P^r_t$ is the baseline, 
while models $V_1$ and $V_3$ are trained with $P^f_t$, without and with temporal losses, respectively. 
$V_3$ gives significantly improved evaluations from the participants. We also outperform DwNet with high confidence, confirming the effectiveness of the temporal losses and the tOF and tLP evaluations. }

\section{More results}
\label{sec:app_more_results}
Similar to \myreffig{fig:more_optimization} in the main paper, we show more examples of $P^o_t$ in \myreffig{fig:app_more_optimization}. It becomes visible that our extrapolation and optimization pipeline can significantly improve the spatial quality of UV coordinates and recover the full silhouette. We also show more virtual try-on applications in \myreffig{fig:more_try_on}, from which we can see that $P^g_t$ generated from our model can be efficiently applied to change clothing texture. 
\you{ For the virtual try-on application, we replace the clothing texture from the original constant texture $T_o$ with a new texture for an updated constant texture $T_o$. Then, the new sequence is synthesized with the updated $T_o$. It is worth pointing out that since we focus on the clothing, we reuse the texture of other parts from the source video so that the evaluation can focus on these regions.
%, which do not interact with the clothing. 
As shown in \myreffig{fig:synthesizing}, the head and feet do not interact with the clothing, so we reuse the texture of those parts from $T_t$ to synthesize $I'_t$.}
\youRevision{
Our pipeline can also be applied to datasets containing more diverse motions and complex backgrounds, such as the Tai-Chi dataset (see results in \myreffig{fig:new_transfer}a).
Results are in line with the conclusions of our main paper: 
our optimization result $P^o_t$ successfully recovers the missing UV coordinates and generates full images $I'_t$. After training, our synthesized result $(P^g_t+T_o)$ is closer to the reference than DensePose $(P^r_t+T_o)$ for temporal and spatial evaluations.
Lastly, our models are specific to garment silhouettes, not individual videos. 
Retraining is only necessary if the 
silhouette changes. E.g. in \myreffig{fig:new_transfer}b, the model is trained with the sequence B (with sleeveless dress) and can be conditioned on poses in A (with long sleeves) to generate C.
We aim for this direction since the silhouettes of common clothes are limited.}

\end{document}